# Bioplausible multiscale filtering in retino-cortical processing as a mechanism in perceptual grouping

Nasim Nematzadeh · David M. W. Powers · Trent W. Lewis

**Abstract** Why does our visual system fail to reconstruct reality, when we look at certain patterns? Where do Geometrical illusions start to emerge in the visual pathway? How far should we take computational models of vision with the same visual ability to detect illusions as we do? This study addresses these questions, by focusing on a specific underlying neural mechanism involved in our visual experiences that affects our final perception. Among many types of visual illusion, 'Geometrical' and, in particular, 'Tilt Illusions' are rather important, being characterized by misperception of geometric patterns involving lines and tiles in combination with contrasting orientation, size or position. Over the last decade, many new neurophysiological experiments have led to new insights as to how, when and where retinal processing takes place, and the encoding nature of the retinal representation that is sent to the cortex for further processing. Based on these neurobiological discoveries, we provide computer simulation evidence from modelling retinal ganglion cells responses to some complex Tilt Illusions, suggesting that the emergence of tilt in these illusions is partially related to the interaction of multiscale visual processing performed in the retina. The output of our low-level filtering model is presented for several types of Tilt Illusion, predicting that the final tilt percept arises from multiple-scale processing of the Differences of Gaussians and the perceptual interaction of foreground and background elements. The model is a variation of classical receptive field implementation for simple cells in early stages of vision with the scales tuned to the object/texture sizes in the pattern. Our results suggest that this model has a high potential in revealing the underlying mechanism connecting low-level filtering approaches to mid- and high-level explanations such as 'Anchoring theory' and 'Perceptual grouping'.

## 1 Introduction

We investigate here whether computational modelling of vision can provide similar interpretation of visual data to our own experiences, based on simple bioplausible modelling of multiscale retinal cell responses to the visual scene. Our visual perception of the world is the result of multiple levels of visual processing. This starts with multiple levels of visual filtering within the retina and ends in multiple levels of processing in the visual cortex. The bottom-up visual processing gives rise to simple percept or features, but multimodal and top-down information flow leads to more complex concepts, as well as influencing basic perception. In our visual system, fast and accurate visual processing functions as a parsimonious system with

N. Nematzadeh (✉) · D. M. W. Powers · T. W. Lewis
College of Science and Engineering, Flinders University,
GPO Box 2100, Adelaide, SA 5001, Australia
e-mail: nasim.nematzadeh@flinders.edu.au
URL: http://www.flinders.edu.au/people/nasim.nematzadeh

D. M. W. Powers
e-mail: david.powers@flinders.edu.au

T. W. Lewis
e-mail: trent.lewis@flinders.edu.au

Springer

minimal redundancy, with the processing in the retina serving different purposes and operating in different ways from the later processing levels of the cortex. We regard this assumption as fundamental to a biologically plausible vision model, and human-competitive computer vision, reflecting our understanding of human vision.

Even given the increasingly detailed biological characterization of both retinal and cortical cells over the last half a century (1960s–2010s), there remains considerable uncertainty, and even some controversy, as to the nature and extent of the encoding of visual information by the retina, and conversely of the subsequent processing and decoding in the cortex (see e.g. the review of physiological retinal findings by Field and Chichilnisky [1] and Golish and Meister [2]).

### 1.1 History of Geometrical illusions

The visual distortion experiences we encounter in visual illusions give clues as to some of the biological characteristics of our visual processing that result in some erroneous perception. The explanations of optical illusions rely on our interpretation of the world, and the ambiguities against our visual experiences result in the illusory percept.

There are various types of optical illusions, and in "Appendix" (Table 1) we illustrate important representatives of the various families including impossible 3D arrangements such as Penrose Triangle and Penrose staircase [3–5], stimuli with multistable perception, flipping back and forth between different perception such as Necker cube [6]; also Herring's and Orbison's illusions [7–9] consist of horizontal and vertical lines located on a part of a radial display, inducing tilt/bow/bulge as a result of the three-dimensional percept and the perspective clues. Hermann Grid and Mach Bands have commonly accepted explanations involving the low-level visual retinal/cortical processing by simple cells [10, 11] and the Lateral Inhibition (LI) mechanism, which some of them need high-level explanations. Table 1 in "Appendix" reflects similarity of illusions, but due to the shortage of space and table arrangement, it may not exactly match other illusion classifications [12] based on other explanations. A complete reference list to the source and original illusory patterns given in Table 1 is also provided in "Appendix" (Table 2). A neurobiological explanation for a variety of Geometrical illusion can be found in [13].

The patterns explored in this paper are 'second-order Tilt' Illusions [14] (Tile Illusions) involving the enhancement of contrast between textural elements of a background such as a checkerboard, for example Café Wall and Bulging checkerboard illusions. In the Café Wall illusion, the illusory tilt percept is the result of mortar lines between shifted rows of black and white tiles. The mortar lines have an intermediate brightness between the brightness of tiles, giving rise to appearance of mortar lines as divergent and convergent instead of parallel lines. On the other hand, in the Bulge patterns, superimposed dots on top of a simple checkerboard give rise to the impression of a bulge or tilt. This is highly affected by the precise position of dots. The illusory perception of these patterns is connected to the figure and ground perception, and in particular, in the Bulge patterns, grouping of dots together creates an illusory figure shape, on top of a textured background (here a checkerboard can be also a grid). This produces apparent border shifts of the checkerboard edges and increases or reduces the impression of the bulges or bows in these patterns.

### 1.2 Simultaneous Brightness/Lightness Illusions

To be able to differentiate between competing explanations for Geometrical illusions, consider the existing techniques for explaining the simultaneous Brightness/Lightness Illusions such as variations of White's effect [15–17]. When investigating Brightness/Lightness Illusions, high-level explanatory models are sometimes seen as a result of lightness shift of the same luminance (thus brightness), which decodes as different lightness for example, in [18]. High-level explanations need higher cortical processing of visual clues such as lightness/transparency as well as past experiences and inferences [19–21]. At the same time, to address their final percept, they might involve ideas of interpolation (1D) or filling-in (2D) [22–25] as well. More recently, the 'Anchoring theory' of Gilchrist et al. [26] is based on 'grouping factors' that signal depth information, without any consideration of the spatial frequency of the pattern. Further explanations for these illusions rely on 'Junction analysis' like T-junctions [27] and 'Scission theory' [28, 29] which triggers the parsing of targets into multiple layers of reflectance, transparency and illumination in which erroneous decomposition leads to Brightness/Lightness Illusions.

Modern 'low-level theories' on Brightness/Lightness Illusions, for example Kingdom and Moulden [30], and Blakeslee and McCourt [17, 31], suggest that a set of spatial frequency filters at early stages of visual processing, mainly preprocessing in the retina are responsible for some Brightness/Lightness Illusions. Recent investigations on diverse range of lightness/brightness/transparency (LBT) illusions by Kingdom [32] have shown different origins for some of these effects due to whether illusion arises from encoding of brightness or the lightness of the pattern. He concluded that the most promising developments in LBT is a model of brightness coding based on 'multiscale filtering' (models such as [17]) in conjunction with 'contrast normalization'.

## 1.3 Tilt illusions

Throughout the history of Geometrical illusion, a variety of low-level to high-level explanations have been proposed covering many 'Tilt Illusion' patterns since Herring's and Helmholtz time as reflected in the recent overview by Ninio [8]. However, there has been little systematic explanation of model predictions of both illusion magnitude and local tilt direction of tilt patterns that reflect subjective reports from the patterns especially on the chosen Tile Illusions. Although Ninio investigated many Tilt Illusion patterns and presented several principles such as Orthogonal Expansion and Convexity Rule, he stated that these explanations cannot address the family of twisted cord [33] and Tile Illusions (investigated here), even though some of these principles might be part of the explanations. There are also other theories for explaining these illusions. For example, Changizi and his team propose a new empirical regularity for systematization of illusions [9], motivated by the theory of 'perceiving-the-present'. This theory is based on the neural lag, which is a latency of 100 ms between retinal stimulus and final perception. The well-known hypothesis of 'perceiving-the-present' [34, 35] has its foundation in the belief that 'the visual system possesses mechanisms for compensating neural delay during forward motion' [9, p. 459] and that we tend to perceive the present rather than perceiving the recent past. Although the hypothesis has been applied for explaining Geometrical illusions in the past [36–39], Changizi's new prediction generalized this idea and categorized Geometrical illusions based on the central idea that 'the classical Geometrical illusions are similar in kind to the projections observers often receive in a fixation when moving through the world' [9, p. 461]. There are some critiques of Changizi's systematization of illusions,[1] such as Brisco's article [40].

However, Tile Illusions have not been explained completely by these so-called generalized theories. Many of these patterns might generally be considered as high-level illusions relying on later cortical processing for their explanation such as Complex Bulge patterns. In the Tile Illusions, the Café Wall illusion [41–47] is the main pattern, which has been investigated broadly. In our previous work [48–50], we have shown that a simple model of multiscale retinal/cortical processing, in the early stages of vision, is able to highlight the emergence of tilt in these patterns with the main focus on the Café Wall pattern.

The current explanation techniques available for investigated patterns are mainly based on three different approaches including the theory of 'Brightness Contrast and Assimilation' mentioned by Jameson [51] and highlighted in Smith et al. [52], 'Perceptual inferences and Junctions analysis' providing high-level explanations (Grossberg and Todorovic [23]; Gilchrist et al. [26]; Anderson and Winawer [29]) and 'Low-level spatial filtering' (Morgan and Moulden [46]; Earle and Maskell [44]; Arai [53]). It is often not obvious how substantive the difference between the explanations is, or whether these explanations are just combinations of orthogonal mechanisms or compatible theories. For example, quite recently, Dixon et al. [54] combined the 'Low-level filtering' explanation of Blakeslee and McCourt [17, 31] with higher-level models such as 'Anchoring Theory' [26], observing that the key idea or common principle in multiscale, inference base and Brightness/Lightness perception is 'high-pass filtering tuned to the object size'.

Based on new biological insights, it is now clear that the retinal output is a stack of multiscale outputs (more details in Sect. 3) and modelling this multilayer representation has a significant power in revealing the underlying structure of the percept in computer vision (CV) models [55–59] such as edges, shades, some textures and even some preliminary cues about the depth information, according to some neurocomputational eye models such as [60, 61]. We adopt a parsimonious approach to modelling vision, in terms of both organizations of complexity and computational cost.

## 1.4 The focus of our investigation

The patterns investigated have some similarities to Brightness/Lightness Illusions such as Irradiation [45, 62], Simultaneous Brightness Contrast (SBC) [63–65] and White's effect [15, 63, 65–69], but the difference between the explanation of these two subclasses of illusion is that for Tilt Illusions we seek for the prediction of tilt not the changes of Brightness/Lightness profile used to describe Brightness induction effects. The patterns under investigation seem to have mid- to high-level perceptual explanations, but in our previous investigations [14, 48–50] we have shown that the illusory tilt cues in these patterns are caused by multiscale retinal/cortical encoding by the simple cells and the Lateral Inhibition among them. We demonstrated how a low-level explanation of these patterns at multiple scales reveals the tilt cues in the local processing of the pattern, followed by higher levels of processing in the retina and the cortex for the integration of the local tilt cues for the final percept. Two samples of Tile

---

[1] Changizi's explanation and systematization of Geometrical illusions not only considers perspective clues, but also examines the probable observer's direction of motion using vanishing point cues, as well as the changes we perceive following the neural lag in the perceived projected size, speed, luminance–contrast, distance and eccentricity of the stimulus which are six correlates of the optic flow, used for illusion prediction and classification.

Illusions are shown in Fig. 1, namely Trampoline [70] and Spiral Café Wall [71] patterns.

In this study, we further explore the neurophysiological model of multiple-scale low-level filtering developed by Nematzadeh et al. [14], based on the circular centre and surround mechanism of classical receptive field (CRF) in the retina. For filtering, a set of the Differences of Gaussians (DoGs) at multiple scales is used to model the multiscale retinal ganglion cell (RGC) responses to the stimulus. The simulation output is an edge map representation at multiple scales for the visual scene/pattern, which utilized to highlight the tilt effects in the investigated patterns here. A systematic prediction of perceptual tilt is presented in [48–50] using Hough space [72] for quantitative measurement of tilt inside the DoG edge map. This multiple-scale representation has some analogy to Marr's and Hildreth [73] suggestion of retinal 'signatures' of the three-dimensional structure from a raw primal sketch, this being supported by physiological evidence [1, 74, 75].

One connection of our model with existing explanations is the concept of assimilation and contrast in perceived brightness. Jameson's dual model of 'Brightness Contrast and Assimilation' [51] explains Brightness/Lightness Illusions in terms of DoG filters with different characteristics and dimensions. Here, the ratio of the filter size to image features results in some brightness shifts, contrast or assimilation. Also this filtering representation at multiple scales might be the underlying mechanism to connect similar explanations such as ours with some mid- to high-level explanations for example 'Anchoring theory' [26] and its extensions such as 'Double-Anchoring' [76] and the idea of illumination framework proposed to address brightness induction effects. Another important outcome of this DoG edge map representation is that it highlights a possible neural mechanism in 'perceptual organizations' for local and global percept, the idea in Gestalt psychology for perceptual grouping of pattern elements. What we mean by 'pattern elements' are smaller elementary components of patterns that lead to the final percept in general and perceiving illusions in particular.

In the next section, we present the psychological view of perceptual grouping and Gestalt psychology in visual perception. The aim is to bridge between low-level spatial frequency filtering (mainly retinal preprocessing) and high-level perceptual organization (Sect. 2). We then move to a detailed examination of the role of multiscale representation in computational models of vision, with a focus on evidence of multiscale filtering within the retina (Sect. 3) contrastively with other models and theories available in prediction of Brightness/Lightness Illusions and Tilt Illusions. Next (Sect. 4), we explain the details of our simple bioplausible Difference of Gaussian (DoG) model, implementing a classical receptive field (CRF) of simple cells in the retina/cortex, in which their scales are tuned to the object size, ending with experimental results. We conclude by highlighting the outcomes, advantages and disadvantages of our simple visual model and proceed to outline a roadmap of our future work (Sect. 5).

## 2 Perceptual grouping

The perceptual view of visual psychology is mainly based on Gestalt psychological findings [77–79]. These are related to the laws about perceiving meaningfully and generating whole forms by the brain as a global figure rather than recognition of its simpler elements such as lines and points. Therefore, the outlook of Gestaltism in perception is conceptually different from the structuralist view and hence criticized by some scholars from computational neuroscience and cognitive psychology. They claim that Gestalt principles can just provide descriptive laws rather than a perceptual processing model [80]. However, the idea of Gestalt psychology attracted many scholars on relevant areas of vision research, which led to many research findings on object recognition and pattern perception in general [81].

Two major reviews of empirical and theoretical contributions of Gestalt psychology in visual perception by Wagemans et al. and Spillmann [77–79] have been released to clarify the meaning and importance of this concept and to bring it under the attention of researchers in the field of vision. The book 'Visual Perception' [79] is among a few that aimed primarily to correlate perceptual phenomena to their underlying neural mechanism. Modern NeoGestaltist views on cortical processing relates to how the brain is working. 'Often the whole is grasped even before the individual parts enter consciousness' [82, p. 10]. This 'arises from continuous global process in the brain, rather than combinations of elementary excitations' [82, p. 11].

There are a few well-known visual theories in the same spirit as Gestalt. One is the 'Reverse Hierarchy theory' [83] claiming that there is a fast feedforward swap that quickly activates global percept in high-level areas with large receptive fields (RFs). There is feedback from these higher areas to lower areas and recurrent processing in the lower-level areas. Processing in lower areas is with small receptive fields for fine-grained processing of local detail of visual input. Another theory is 'General Theory of Visual Object Recognition' [84], in which two streams of processing occur in parallel: high spatial frequencies of input images which are processed relatively slow, and in a feedforward sequence in the visual cortex (V1, V2, V4 and so forth), and, on the other hand, low special frequencies of visual input, which are quickly transmitted to area in prefrontal cortex to identify the object as well as their most likely scene context. These two streams are integrated and refined in an interactive, reiterant way for the final percept. Both of these theories postulate that global processing comes first, and they have both dynamic views on cortical processing.

In this context, Spillmann et al. state that 'Over the years, theoretical accounts for RF properties have progressively shifted from classic bottom-up processing towards contextual processing with top-down and horizontal modulation contributing. These later effects provide evidence for long-range interaction between neurons relevant to figure–ground segregation and pup out by brightness, color, orientation, texture, motion, and depth' [85, p. 1]. Specific models for uniform surfaces, filling-in and grouping [86, 87] have been formulated and tested to enable the transition from local to global processing by using information from the beyond the classical RFs [85]. It has also been shown that Gestalt factor of good continuation is critical for contour integration [88].

It has been proven that neuronal response not only depends on local stimulus analysis within the classical RFs, but also from global feature integration as a contextual influence, in which it can extend over relatively large regions of the visual field [89]. This is another evidence for the Gestalt credo that a whole is not reducible to the sum of its parts [85]. 'Classical RFs increase in size from near foveal to peripheral location, from V1 to higher areas in the extrastriate cortex. Smallest in the primary visual cortex (V1), larger in V2, larger again in V3A and V4. Also the slope of the functions describing the increase in size with eccentricity increase progressively from lower to higher visual area' [85, p. 7].

The Gestalt principles of object and element perception are about grouping of objects based on their similarity, proximity or other cues. Within this global perception processing, there are some innate mental laws reviewed in Wagemans et al. [78]. We believe that in general, Gestalt grouping laws of 'closure', 'proximity', 'similarity' and 'continuity' are among the principles, which their underlying neural mechanism can be revealed by some extent by low-level vision models. The simple implementation of retinal/cortical multiscale encoding in our model [14, 48–50] provides some basic understandings of these Gestalt principles, which will be explained further for investigated patterns in Sect. 4.2.

The aim here is to connect multilevel explanations from perceptual organization with global and local percept, to the background low-level filtering explanations. There is a similar connection of low-level retinal/cortical processing to high-level Gestalt grouping principles stated by other researchers in the field. For instance, Craft et al. [90] propose a neural mechanism of 'figure–ground organization' based on border ownership to model complex cells in the primary visual cortex (V1). In their investigation of grouping cell connections, they linked their findings to Gestalt grouping principles such as 'connectivity', and 'convexity' by applying connection weights based on different sizes of receptive fields in the cortex which was modelled based on multiscale and orientated DoG filters. Similarly, Roelfsema's [91] findings in the perception of 'pathfinder' suggest that the Gestalt principle of 'good continuation' can be understood in terms of the anatomical and functional structure of the visual cortex.

We have shown [14, 48–50] that the retinal/cortical simple cells processing, simulated in this model for these stimuli (Tile Illusions), explains the emergence of tilt in these patterns. It also provides a basic connection to perceptual grouping of pattern elements and their relational organizations by simple modelling of classical receptive fields (CRFs), tuned to the object size. We will show in Sect. 4.2 how a specific group of pattern elements can be generated in the edge map based on different parameters of our bioderived DoG-based model, simulating the responses of simple cells in the retina/cortex to a stimulus.

We close this section with the following key points by Wagemans who asserts that 'True Gestalts are dynamic structures in experience and determine what will be wholes and parts, figure and background' [82, p. 18]. 'In fact, Gestalt phenomena are still not very well integrated into mainstream thinking about the visual system operating principles (e.g. selectivity and tuning of single cells, V1 as a bank of filters or channels, increasing receptive field size and invariance at higher levels of the hierarchy)' [82, p. 18].

## 3 Multiscale retinal/cortical representation/biology and modelling

Hubel and Wiesel [92] showed that our visual system has a multiscale and orientation filtering mechanism referred to as an orientation tuner with columnar representation of cortical cells. In contrast, other classical investigations such as Barlow [93] and Kuffler [94] suggested that the image encoding in the retina is based on a centre–surround organization mechanism in retinal successive layers also known in cortex.

Although physiological evidence of the existence of multiscale filtering in the receptive fields of the retina is well known, we highlight the changing size of Ganglion cells (GCs) with eccentricity (that is their distance from the fovea). A recent biological study [1] indicates there are at least 17 distinct GCs in the retina. Each type has a diverse range of size in relation to the eccentricity of neurones and the distance from the fovea. The layer ganglia, covering the retina, thus include different sizes and scale receptive neurons, resulting in a multiscale retinal encoding of the visual input that will be sent to the cortex. This is our assumption for low-level visual processing in the retina. There are other factors strengthening this assumption, such as 'Fixational eye movements' [75] which are critical to prevent fading the whole visual world. So the eyes are never still and even when we are fixating, there is a visual mechanism of continuously shifting retinal image on the retina by factor of a few 10–100 s GCs due to the type of fixational eye movements. These include microsaccades, tremors and drifts, which are unconscious source of eye movements. There are also conscious eye movements such as saccades that result on different retinal cells sensations. All the above-mentioned evidence indicates the multiscale filtering representation in the retina due to the eccentricity relation of the GCs sizes, which is their distance from the fovea.

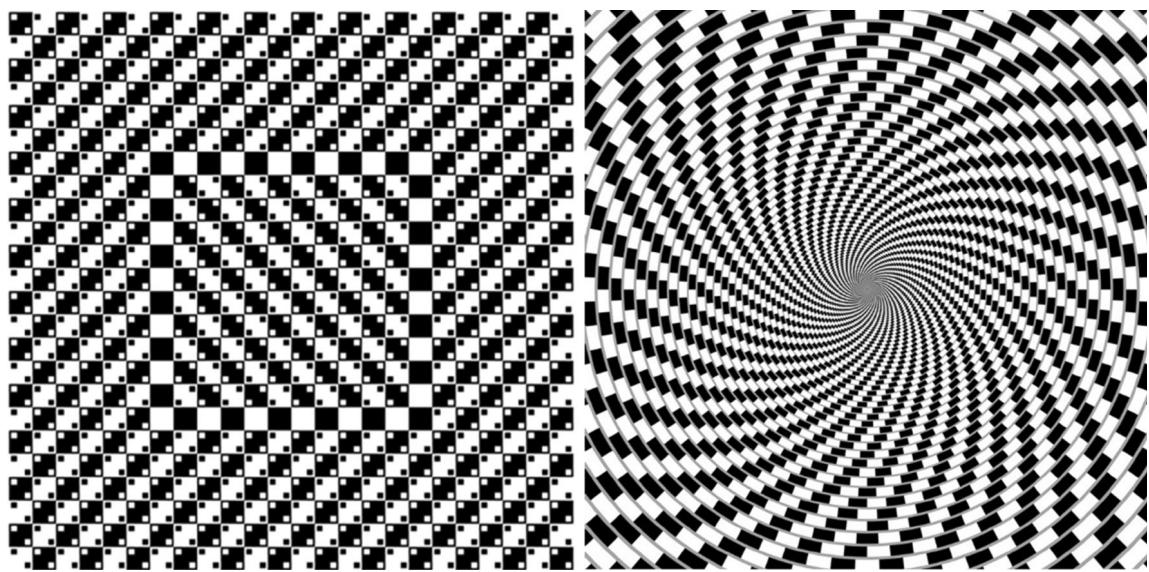

**Fig. 1** Sample Tile Illusion patterns. *Left* Trampoline pattern [70], *right* Spiral Café Wall illusion [71]

Beyond the multiscale nature of retinal representation, it has also been shown that some retinal receptive fields have far extended surround seen in the lateral mechanism of both horizontals and amacrine cells in the retina compared to the retinal classical receptive fields (CRFs) size. The idea of non-classical receptive fields (nCRFs) was first introduced by Passaglia et al. [95] with modelling RFs based on widely spread surround. This representation provides some evidence for even specific orientation tuning cells in the retina. Therefore, retinal visual output contains not only the edge map with multiscale edge information including the shades and brightness profile around the edges, but also their orientations as a result of multiscale and orientation processing of some retinal cells. This along with other evidence suggests that in many stages of our visual processing, there is spatial filtering adjustment involved such as eye movements, which creates an adaptable spatial size mechanism in the retinal ganglion cells [54]. Shapiro and Lu [64] argue that for brightness perception, the spatial filter size is relative to the object size in the whole image, which is consistent with the spatial vision literature not the retinal spatial frequency. Our DoG model has the same property.

The implementation of the receptive fields (RFs) in both the retina and the cortex based on the Differences of Gaussians (DoGs) dates back to 1960s when Rodieck and Stone [96] and Enroth-Cugell and Robson [97] showed an efficient model for retinal ganglion cell responses and the centre–surround antagonistic effect. The explanations for some Geometrical illusions such as Mach Bands, SBC and Hermann Grid rely on the 'Lateral Inhibition' and 'contrast sensitivity' of the RGCs [98, 99]. A model of ontogenesis of lateral interaction functions was derived mathematically by Powers [100] showing theoretically that it was possible for commonly assumed types of lateral interaction function to self-organize and in particular approximate the DoG and LoG models, as well as other models related to the Poisson and Gaussian distributions. Considering how this bootstraps to a higher-level model, such distributions can then explain the repeated patterns of edge detectors at particular angles through a self-organizing model [101]. Our investigation is based on applying a low-level DoG filtering model, simulating the responses of RGCs to the Tile Illusion stimulus, to find a lateral inhibitory explanation for these illusions.

Despite the controversy regarding low-level explanations for simultaneous Brightness/Lightness Illusions for example [98, 102–104], low-level filtering techniques showed their great power in addressing illusions of type Brightness/Lightness inductions for example [31, 67, 105, 106]. Some of these models build up their methods on nCRFs implementation such as [17, 68] by using elongated surround for orientation selectivity of some retinal RF or cortical cells. Tile Illusions may have some similarities to Brightness/Lightness Illusions, but they are usually given different explanations [42, 45, 107–110]. We clarify that for Tilt Illusions we should predict and measure the tilt orientation not the brightness profile changes, which is needed for brightness induction explanations, although analysis of brightness might provide us further clues having indirect effect on tilt.

In our previous investigations [48–50], we have shown that a simple classical receptive field model, implementing multiscale responses of a symmetrical ON-centre and OFF-surround RGCs, can easily reveal the emergence of tilt in these patterns. In the future, we intend to extend the model to nonlinear spatial subunits and/or a nCRF implementation in order to identify angles of orientation on detected tilts in the edge map quantitatively instead of our Hough analysis stage.

The novelty of this work arises from its simplicity and the multiple-scale representation view, to explain the emergence of tilt in Tile Illusions based on a DoG edge map of the visual stimulus, in which the scales of filters are determined based on the characteristics of the investigated pattern. More importantly, we believe that the edge map explanation for the tilt effect and visual perception in general bridges the low-level multiscale filtering with high-level explanations of Gestalt perceptual grouping structures by principles such as good continuation and connectivity of the pattern elements. Furthermore, we should state here that gradual changes of DoG scales in here make the model more biologically plausible, and abrupt changes like doubling the scale in each level might end up losing some important information, for modelling CV tools in general and addressing Tilt Illusions.

## 4 Our model

Biological evidence has shown that GC excitation has a centre–surround organization [111] and could be modelled by the differences of two Gaussians [112]. Marr and Hildreth [73] suggest for an involvement of higher-order Gaussian derivatives, utilizing the Laplacian of Gaussian (LoG) and its DoG approximation to model the initial retinal filtering. Young [113, 114] applied linear combination of Gaussians and LoG instead of DoG, but there is still no biological evidence for the structure of these functions [115].

A neurophysiological inspired model implementing the lateral inhibition by the retinal cells [96, 97, 116] is employed by Nematzadeh et al. [14, 48] based on the DoG filtering at multiple scales simulating retinal RF responses to address Tilt Illusions in general. The output of the model is a DoG edge map at multiple scales, in which each scale of the DoG creates a new layer of visual information. Our main intention here is to connect the edge map representation of our model (simulating retinal GC responses to stimuli) to higher-level perceptual grouping mechanisms such as Gestalt grouping principles.

### 4.1 Multiscale implementation of Difference of Gaussians

The interaction of lateral inhibition by retinal GCs can be modelled by differences of two Gaussians, one for the centre and one for the surround, in which the surround has a larger standard deviation compared to the centre. The DoG filtering techniques are generally used for identifying the edges in CV applications, and by additional multiple-scale DoG filtering, an edge map representation for the pattern is extracted. The multiscale Gaussian representations used in diverse range of CV applications such as DoG/LoG pyramids [58], SURF [117, 118] and SIFT [119] for image representation design, by considering the trade-off between efficiency and complexity for both rendering smooth regions and detailed contours and textures in the image [59]. Due to the bioplausibility goal, we have considered just a simple implementation of multiple-scale DoG filtering and have not used any of those above-mentioned CV techniques to prevent any alternation of the results. Also we change the DoG scales as incremental/decremental stepwise in each level, to capture the maximum output results as the levels of edge map, instead of multiplying the scale in each step, which is mainly used in the CV models.

For a pattern $I$, the DoG output of a retinal GCs model with circular centre and surround organization is calculated as follows:

$$\text{DoG}_{\sigma, s\sigma}(x, y) = I \times 1/2\pi\sigma^2 \exp[-(x^2 + y^2)/(2\sigma^2)] \\ - I \times 1/2\pi(s\sigma)^2 \exp[-(x^2 + y^2)/(2s^2\sigma^2)] \quad (1)$$

where $DoG$ is the convolved DoG result, $x$ and $y$ indicate the distance from the origin in the horizontal and vertical axes, respectively, $\sigma$ is the sigma or 'scale' of the centre Gaussian ($\sigma = \sigma_c$) and $s\sigma$ shows the scale of the surround Gaussian ($s\sigma = \sigma$). Therefore, $s$ factor is referred to as the *Surround ratio* as shown in Eq. (2). In vision models, this factor is sometimes referred to as point spread function of retinal cells [120].

$$s = \sigma_{\text{surround}}/\sigma_{\text{center}} = \sigma_s/\sigma_c \quad (2)$$

The value of $s$ in our model set to 2. Other values, such as 1.6, were tested with negligible difference in the result.

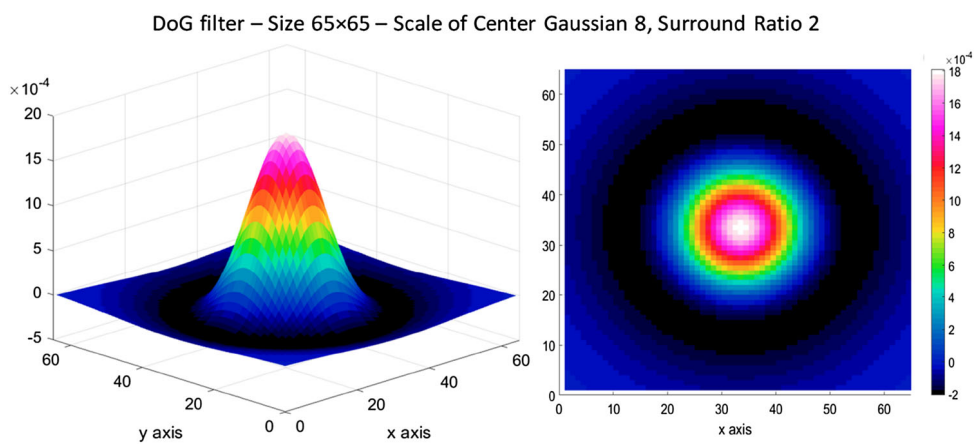

**Fig. 2** *Left* 3D surface of a Difference of Gaussian filter with the scale of the centre Gaussian equal to 8 ($\sigma_c = 8$). The Surround ratio is $s = 2$, and the Window ratio is $h = 8$. *Right* Top view (2D) of the DoG filter. Window size is $65 \times 65$. Jet white colour map is used to display the graph. (Color figure online)

Increasing the *s* factor results in more surround suppression effect on the final output, while the local weights are decreasing (due to definition of a normalized Gaussian kernel).

Rather than the *s* factor for the surround Gaussian, the DoG window size is also another parameter to consider, characterized by parameter *h* given in Eq. (3). Very large windows result in long computation, and very small windows are just approximating a box blur filter not a weighted Gaussian one. For the experimental results (Sect. 4.2), the *h* value set to 8 times larger than the scale of the centre Gaussian ($\sigma_c$) to capture the inhibition effect as well as the excitation and keep the significant values of both Gaussians within the window (95% of the surround Gaussian is included within the DoG filter). A three-dimensional surface graph and a top view of a sample DoG filter are shown in Fig. 2 in jet white colour map.[2]

Window size = $h \times \sigma_c + 1$ (3)

So, to generate a multiple-scale DoG representation for the pattern we apply DoG filter at different scales to the pattern. The scales are pattern specific. The outputs of the DoG model, as the edge maps of sample patterns, are displayed in Figs. 3, 4, 5 and 6. On top of these figures, you see the original patterns (Café Wall, Munsterberg, Crop of Café Wall and Complex Bulge patterns) and an enlarged DoG at a specific scale from the edge maps that highlights the tilt effect in these patterns. The edge map representations at multiple scales (may be referred to as multiple-scale DoG edge map for convenience in our work) have been presented at the bottom half of these figures for six or eight scales (six or eight levels of $\sigma_c$). Rather than the scales of the centre Gaussian, the other parameters of the model and the characteristics of each pattern have been provided in each figure. In the edge map output, the scale of the centre Gaussian ($\sigma_c$) increases first from left to right within a row and then from one row to the next. We tried to represent the multiple-scale representation of our bioplausible low-level filtering model, in a way that the output result can be seen easily as a sequence of increasing scales, which in total provides us the multiple-scale edge map representation for each pattern.

The result of our computational model interestingly reveals the possible underlying neural activation/inhibition effect of the retinal ganglion cell (RGC) responses in our simulations to these illusion stimuli (Tile Illusions). As the results in Figs. 3, 4, 5 and 6 show, the edge map representation of these patterns connect to some 'Gestalt perceptual grouping' principles explaining high-level perceptual organization of pattern elements such as 'proximity', 'similarity', 'continuity' and 'good continuation'. These grouping principles are assumed to be the basic blocks for our perception of the world.

Based on the experimental results provided in Figs. 3, 4, 5 and 6, we found that different grouping structures of pattern elements emerge in the gradual change of scale in the edge map. The grouping structure in the DoG edge map at multiple scales is related to factors such as scale resolution including number of scales and increment/decrement of scale level, the size of DoG filters, as well as the predefined scales in the model which determined by the spatial frequencies of the objects/texture sizes in an investigated pattern which is consistent with spatial vision literatures.

In physiological vision, it is more likely that there is scale adjustment by a preprocessing retinal mechanism or a physical mechanism such as fixational eye movements [75]. This would seem to optimize in some sense for the later retinal processing and the encoding of its multiscale representation for further analysis in the cortex and would involve feedback mechanisms, which provide local supervision in what is overall an unsupervised system, again a matter for further research. This scale adjustment could be modelled in CV by pre-analysis of spatial frequency of the pattern's features before setting up the scale values for the bioderived DoG model, to produce a parsimonious efficient representation out of the pattern. We have defined the scales of the DoG filters empirically, by considering the size of elements inside each pattern.

---

[2] Downloadable from Mathworks—MATLAB central. http://www.mathworks.com/matlabcentral/fileexchange/48419-jetwhite-colours-/content/jetwhite.m.

Therefore, there is a connection between psychological aspects of a pattern and its structural features (considering the grouping factors of its elements) and a parametric low-level representation of the pattern (as its retinal encoding) implemented in our model. So, by modelling the RGC responses and generating an edge map representation of a pattern, we are able to find a precise given range of DoG scales, which reveals the emergence of a particular perceptual grouping structure from the pattern elements. We will show this further in the following results.

### 4.2 Experimental results

The output of our bioplausible model [14, 50] for Tile Illusions, including the 'Café Wall', 'Munsterberg' and Complex Bulge' patterns, shows that simply utilizing a DoG processing at multiple scales, implementing lateral inhibition in the RGCs, not only produces an edge map when applying a fine-scale DoG, but also extracts other hidden information such as shades and shadows around the edges, and even the local texture information by increasing the scale of the DoG filter.

These results not only highlight the connection of our model to Jameson's theory of 'Brightness Contrast and Assimilation' [51], but also indicate that there are numerous Geometrical cues available that can be uncovered by this simple edge map representation.

There are some previous explanations for connecting 'Brightness/Lightness' Illusions and 'Geometrical' illusions [46, 52]. The tilt perception in Tile Illusions seems to be affected from 'Brightness Contrast and Assimilation' as well as some 'border shifts' [14, 45].

The multiple-scale DoG edge map for the 'Café Wall' illusion indicates the appearance/emergence of divergence and convergence of the mortar lines in the pattern shown in Fig. 3, similar to how it is perceived. The Café Wall illusion originates from the twisted cord [33] elements with the local inducing tilt and then integration of these local tilts to an extended continuous contour along the whole mortar line [47, 121]. The investigated pattern is a Café Wall of $3 \times 8$ tiles with $200 \times 200$ px tiles and 8 px mortar. We start at finest scale below the mortar size at scale 4 ($\sigma_c = 4$) and extend the scales till scale 24 ($\sigma_c = 24$) here, with incremental steps of 4. At scale 28 ($\sigma_c = 28$) which investigated but has not been shown in the figure, the filter captures nearly the whole tiles of the pattern [refer to Eq. (3)]. Non-critical parameters of the model are $s = 2$ for the *Surround ratio* and $h = 8$ for the *Window ratio*. Similar explanations for the Café Wall illusion have been given by other scholars based on either 'Low-level filtering' models [41, 43, 44, 46, 110] or 'higher-level' psychological explanations such as 'Border Locking' explanation [42]. The illusion percept in the pattern could be affected by the intermediate brightness profile of the mortar lines with respect to the brightness of the tiles, the height of the mortar lines, the ratio of the mortar height to the tile size, the amount of tile shifts in consecutive rows (phase of tile displacement) [41, 42] and even more involvement of other perceptual parameters in the pattern. The simulation results of our investigations on variations of Café Wall patterns having different characteristics based on the above-mentioned parameters have been presented in our article [122]. The paper presents both qualitative and quantitative comparison results of the tilt prediction of the model for both the magnitude and the direction of tilt in these variations. We note that even the strength of illusion in different variations of the Café Wall pattern is predictable from the DoG edge map representation at multiple scales. We are not aware of any other theories which have produced quantitative predictions for this illusion.

A closer look at the multiple-scale representation of the Café Wall pattern in Fig. 3 reveals further cues about how perceptual grouping structures are generated during the DoG processing. The grouping of white tiles in two adjacent rows by the mortar line connecting them starts to appear at fine scales at 4 and 8 ($\sigma_c = 4, 8$). As the scale increases, this grouping structure (mortar line connection with the tiles), referred to as 'twisted cord elements' [33] of the pattern, starts to get disconnected, as shown at scales 12 and 16 ($\sigma_c = 12, 16$) clearly. At the next scale ($\sigma_c = 20$), there is no grouping visible of the tiles and mortar cues, although the appearance of tiles as wedge shape provides a cue to support the near-horizontally diverging and converging perception of the location of the mortar lines, even without the appearance of them in the DoG output at coarse scales. From scale 20 ($\sigma_c = 20$) onwards, a new grouping arrangement arises, this time in the vertical direction. This group arrangement connects identically coloured tiles in a vertical zigzag direction exactly in the 'opposite direction' to the previously seen (near-horizontal) diverging/converging mortar groupings.

The 'vertical zigzag groupings' seem to be the result of our global perception of the pattern, while the near-horizontal diverging/converging' mortar lines are a local percept arising from a local focus on the mortar lines joining different tiles in adjacent rows. Therefore, there is a simultaneous perception of both groupings due to rapid changes in the focus point of the eye. Even at a focal view, due to eccentricity relation of RGCs sizes and distance to the fovea, we still have this multiscale retinal encoding of the pattern. Although we might get an impression of either the 'local or global percept' of the pattern in an instance of time, it is more likely that the global percept would carry more weight in the final perception of the illusion pattern, but interestingly, even at coarse scales, where tiles are not connected through the mortar lines, the wedge shape of tiles provides the same stable direction for the divergence/convergence percept of the mortar lines in the DoG edge map.

The edge map representation of the Café Wall illusion indicates that, at different scales of the DoG, there are 'incompatible grouping precepts'. In other words, according to 'continuity' principle at scale 8 ($\sigma_c = 8$), we see a column where the middle element is displaced to the right, while at scale 20 ($\sigma_c = 20$) it is displaced to the left. We claim that the existence of different precepts at different scales contributes to the tilt induction in the Café Wall illusion [14, 48].

To further investigate the contribution of the mortar lines on the tilt effect of the Café Wall pattern, we investigated the 'Munsterberg illusion' [41, 46, 47], a version of Café Wall pattern without any mortar lines in between rows of tiles, as shown in Fig. 4. The tile sizes are the same as the Café Wall pattern explained before (200 × 200 px) and the parameters of the model have been kept the same as described in Fig. 3. The Munsterberg pattern and the DoG output at scale 8 ($\sigma_c = 8$) have been shown on top, followed by the multiple-scale edge map of the pattern at six different scales at the bottom half of the figure.

The edge map of the Munsterberg pattern (Fig. 4) shows that the early grouping of nearly horizontal diverging and converging tilt cues along the mortar lines seen in the Café Wall pattern at fine scales ($\sigma_c = 4, 8, 12$; Fig. 3) does not occur with the Munsterberg pattern at all. The only grouping of pattern elements revealed in multiple scales of the DoG edge map here is just the vertical zigzag or columnar groupings of identically coloured tiles; therefore and in direct contrast to the 'Café Wall' illusion, the 'Munsterberg' pattern has NO illusory perception of tilt. The results support previous psychophysical findings [41, 46, 47].

So, the DoG edge map of the 'Café Wall' qualitatively reveals the tilt in the pattern. For quantitative analysis of tilt, we have used Hough space to measure the absolute mean tilt angles of the mortar cues in the edge map of the Café Wall pattern at multiple scales [48–50]. Figure 5 shows the tilt analysis results of a crop section of a Café Wall pattern. The crop section consists of a 4 × 5 tiles and selected from a Café Wall of 9 × 14 tiles with 200 × 200 px tiles and 8 px mortar as shown in Fig. 5 (top). At the bottom left, the edge map at six different scales ($\sigma_c = 8$ to 28) with incremental steps of 4 is displayed (in jet white colour map). The bottom right shows the detected houghlines in green, displayed on the binary edge map at six different scales and around four reference orientations of horizontal, vertical and diagonals (blue lines indicate the longest lines detected at each scale, and red and yellow crosses show the beginning and end of detected line segments). For further implementation details about the quantitative measurement of the degree of tilt in the pattern and detected houghlines, please refer to [48, 50].

The other investigated pattern here is 'Complex Bulge' pattern shown in Fig. 6. The pattern consists of a checkerboard background and superimposed dots on top of the checkerboard. The edge map of the pattern at eight scales has been shown at the bottom half of the figure, starting at scale 1 ($\sigma_c = 1$), the finest scale, to scale 8 ($\sigma_c = 8$), with incremental steps of 1. The DoG output at scale 2 ($\sigma_c = 2$) has been enlarged and displayed on top of the figure, next to the original pattern. Here the *Surround ratio* is chosen as $s = 1.6$ [73], and the *Window ratio* is $h = 8$. As explained in Sect. 4.1, the 'distributions of spatial scales' are specified by pattern elements. In Fig. 6, the size of the Complex Bulge pattern is 574 × 572 px. The dimensions of each individual tile and small dot in the pattern are 36 px and 10 px, respectively. Therefore, in order to capture both high-frequency details (superimposed dots) and low-frequency contents (tiles) from the pattern [using Eq. (3) and the constant value of 8 for the *Window ratio*], the range of scale should start with a filter smaller than the dots and extended to a maximum size larger than the tiles in the pattern (*Window size* = $8 \times 1 + 1 = 9 < 10$ px; the size of dots; and *Window size* = $8 \times 8 + 1 = 65$ nearly twice as the size of each tile to specify the range of scales between 1 and 8 with incremental steps of 1 in the edge map of this pattern).

The bulge effect might be addressed by 'high-level' explanations such as uncertainties in both formation and processing of image features, for instance points and lines [107] or even 'psychological' explanations based on categorization of edges due to different intensity arrangements around them [42, 71], but our explanation relies on a 'low-level multiscale filtering model to address the tilt/bulge effects in these patterns. Based on our assumptions, the 'Bulge effect' is happening due to a number of visual clues, for instance the brightness perception of the checkerboard background causing a simple border shift outwards for the white tiles, and expansions in the intersection angles due to Brightness Contrast and Assimilation theory [51]. More importantly, further clues related to local positions of superimposed dots, which may have frequency discharge or emission results in local border tilts or bows [14].

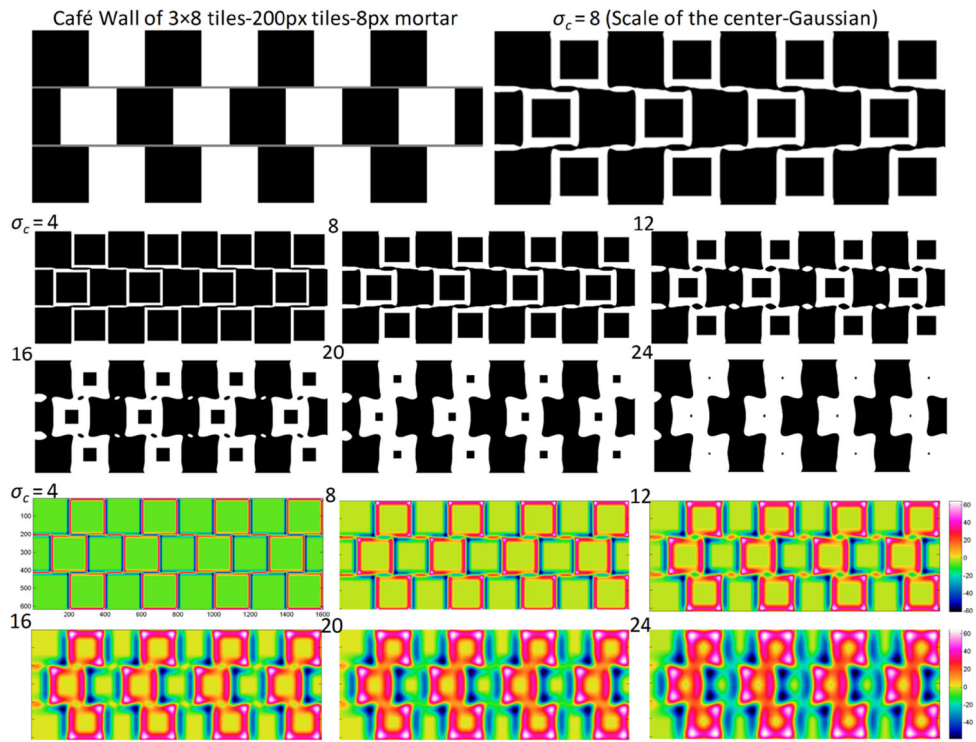

**Fig. 3** *Top left* Café Wall pattern with 200 × 200 px tiles and 8 px mortar. *Top right* Enlarged DoG output at scale 8 ($\sigma_c = 8$) in the edge map. *Centre* The binary edge map at six different scales ($\sigma_c = 4$ to 24 with incremental steps of 4). *Bottom* Jet white colour map of the above edge map. Rather than the centre Gaussian, other parameters of the model are: $s = 2$, and $h = 8$ (the Surround and Window ratios, respectively) (Reproduced by permission from [145]). (Color figure online)

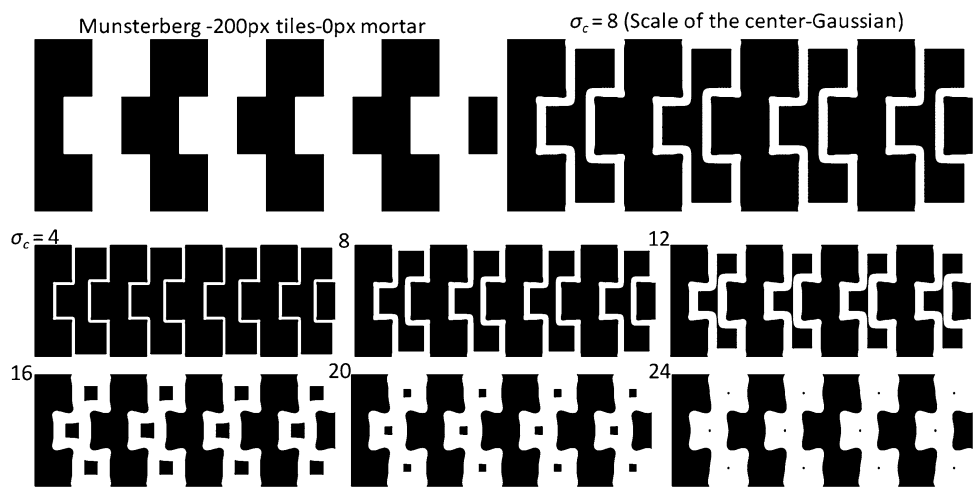

**Fig. 4** *Top left* Munsterberg pattern with 200 × 200 px tiles. *Top right* Enlarged DoG output at scale 8 ($\sigma_c = 8$) in the edge map. *Bottom* The binary edge map at six different scales ($\sigma_c = 4$ to 24) with incremental steps of 4. Other parameters of the model are: $s = 2$ and $h = 8$ similar to Fig. 3 (Reproduced by permission from [145])

In the 'Complex Bulge' pattern (Fig. 6), the output of our DoG model again predicts two distinct groupings. At fine scales from 1 to 3 ($\sigma_c = 1$ to 3), there is a grouping of a two-dimensional bulge, starting from the central tile square, expanding towards the outer regions, like a circular movement of a bulge from the centre to its surroundings.

The effect produces a kind of edge displacement or tilt of the checkerboard edges with the induction of the bulge. The DoG output at fine scales ($\sigma_c = 1$ to 3) in Fig. 6 highlights a grouping of connected superimposed small dots together, which has a close connection to 'Similarity', 'Continuity' and 'Connectivity' in the Gestalt grouping principles. At fine scales of the DoG edge map, the 'central tile' plays an important role in the inducing bulge effect. By increasing the scale from 4 to 8 ($\sigma_c = 4$ to 8), another grouping structure starts to emerge, out of identically coloured tiles with an X-shape structure combined with a wave-like effect between the intersections of the X. Unlike the first bulge grouping, which occurred as a result of fine-scale DoG filtering, the DoG output of the pattern at coarse scales results in edge movements of the 'central tile' with the tilt effect appearing in a different direction (nearly 45° rotated from the grid) due to the 'X-shape grouping' structure, which groups the background checkerboard tiles together. Again, 'Similarity', 'Connectivity' and 'Continuity' are the Gestalt grouping principles that have been shown by our simulated results modelling low-level processing of the simple cells in response to the pattern. The difference between the structure of the X-shape group and the Bulge group is that the X-shape grouping occurs for coarse or low-frequency components of the pattern, that is the checkerboard tiles, rather than the high-frequency details, that is the small superimposed dots (appear in the DoGs at fine scales) in which it generates the Bulge group. The X-shape grouping has an inducing effect of shrinkage on the central tile exactly opposite to its previous expansions with the DoG responses at fine scales.

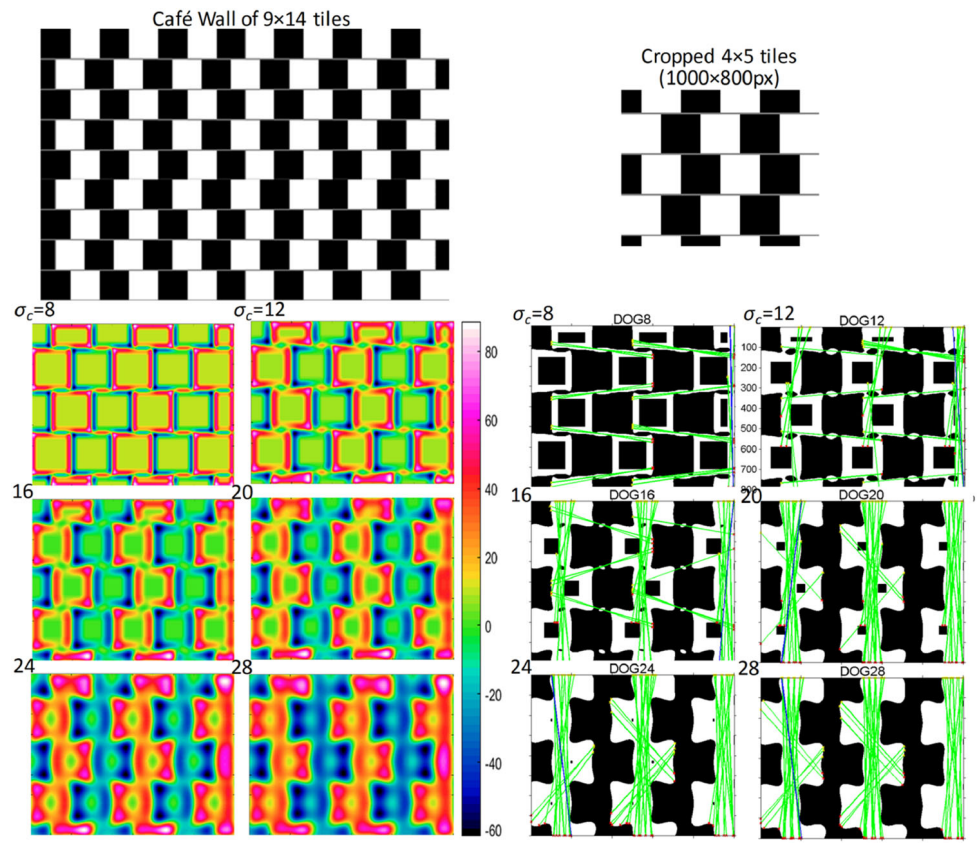

Fig. 5 *Top* A crop section of 4 × 5 tiles (enlarged) from a Café Wall of 9 × 14 tiles with 200 × 200 px tiles and 8 px mortar. *Bottom left* Edge map of the crop section at six scales ($\sigma_c = 8$ to 28), with incremental steps of 4 in jet white colour map. *Bottom right* Detected houghlines displayed in *green* on the binary edge map and around four reference orientations of *horizontal*, *vertical* and *diagonals*. *Blue lines* indicate the longest detected lines at each scale of the edge map (Reproduced by permission from [145]). (Color figure online)

As into the 'Café Wall' illusion explained before, in the 'Complex Bulge' pattern, the DoG output of the model includes two distinct grouping arrangements, in this case a 'Central Bulge' and an 'X-shape' induction, with incompatible effects on the border shifts. It seems that the 'X-shape' percept arises from global perception when we focus on the periphery of the pattern, while the 'Bulge effect' is the result of focusing on the central area close to the central tile or on the inducing superimposed dot cues on the checkerboard; therefore, it is a local percept. To summarize, the multiple-scale DoG edge map of the 'Complex Bulge' pattern reveals two distinct and incompatible precepts or grouping structures arising simultaneously and results from our local to global percept of the pattern, contributing to bulge induction in the pattern.

Tilt effect in the Bulge patterns can also be explained in terms of perceptual interaction of foreground and background elements. The checkerboard in these patterns plays as a background and small superimposed dots as the foreground object. Due to the blurring effect of multiscale retinal GCs encoding and multiscale cortical processing of retinal output, these superimposed dots get connected (grouped) together and create a foreground subpattern as an illusory figure. The final perception in these patterns arises

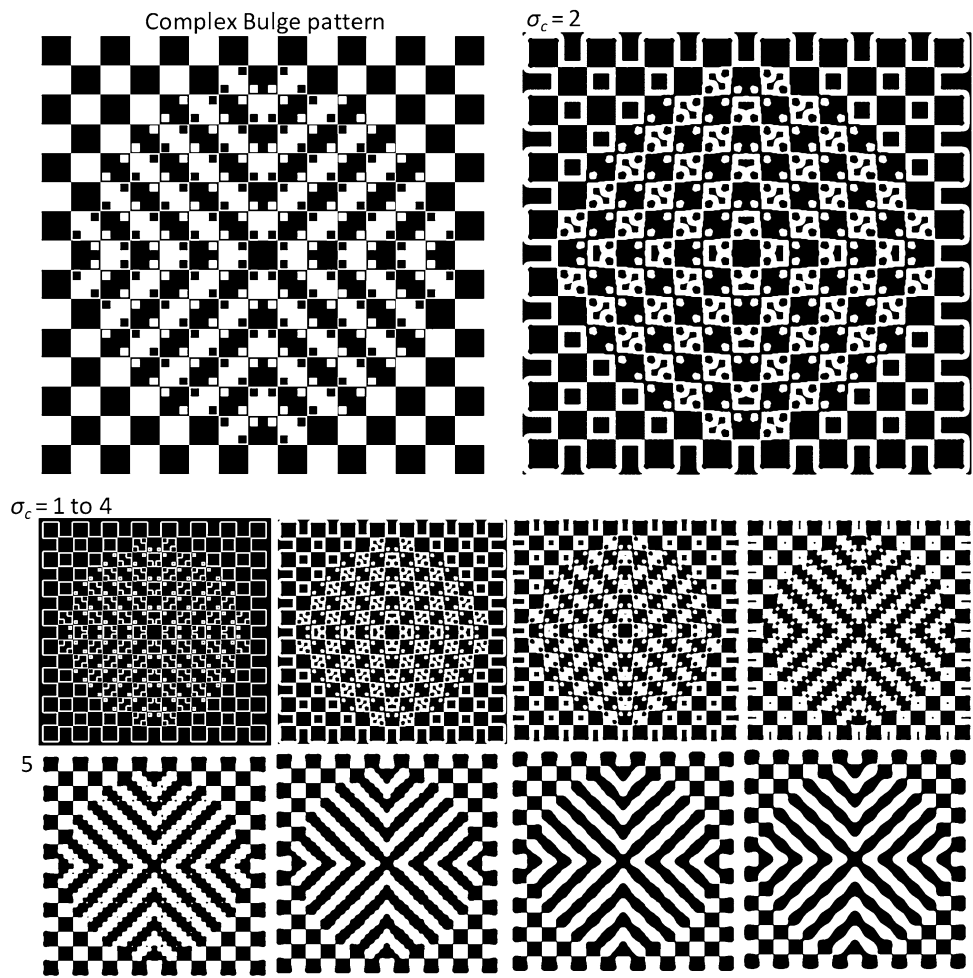

**Fig. 6** *Top left* Complex Bulge pattern (574 × 572 px) and *top right* The DoG output at scale 2 ($\sigma_c = 2$)—enlarged that highlights the bulge effect in the pattern. *Bottom* The binary edge map of the pattern at eight different scales ($\sigma_c = 1$ to 8) with incremental steps of 1. Other parameters of the DoG model are: $s = 1.6$, and $h = 8$ (the Surround and Window ratios, respectively) (Reproduced by permission from [145])

from the interaction of foreground superimposed dots (illusory figure of grouped dots) and the background elements of checkerboard tiles. In our implementation, the blurring effect on the final perception of the pattern is related to the size of the DoG filters, which acts as a band-pass filter. It is noteworthy that the shape of dots does not play an important role in the final percept of these patterns, although their sizes and positions (their relative distance to the outlines of the tiles of the checkerboard) in the patterns are important. They can be square, circle or any other shape but as long as their relative sizes are the same, the overall impression of tilt would be similar. The reason is that in the multiple-scale DoG processing, which simulates visual blurring at early stages of visual processing, the outputs get very similar when the size of DoG filter reaches to the overall size of the dots.

Rather than just showing the bulge/tilt effect qualitatively in this pattern, we have done some quantitative measurement of tilt in Hough space. Detected houghlines for two scales of 2 and 4 ($\sigma_c = 2, 4$) of the DoG edge map have been displayed in green on the binary edge map in Fig. 7 (centre row). The two images at the bottom row of the figure show the zoomed-in versions of the images in the centre row. Here the technique used for quantitative tilt measurement highlights two different groupings of pattern elements at different scales of the edge map. We have shown this quantitative tilt results and detected houghlines on the Complex Bulge pattern here as a sample to show the coverage of our model for more complex Tile Illusions. Further details about the extraction of houghlines on this pattern are out of the scope of this paper, but for Café Wall pattern it can be found in [48–50].

We should note that although the DoG edge map representation of Tile Illusions reveals the local tilt cues at multiple scales of the edge map (most probably reflects on the encoding of the retinal ganglion cells), these local tilt cues must be integrated at higher levels of visual processing (completed in the cortex). The mechanisms involved in the edge integration are widely believed to be the result of cortical complex cells (see, for example, [47, 110, 121] for the Café Wall illusion). Based on the neurophysiological evidence for the existence of some retinal GCs with orientation selectivity property [93, 123–125], as well as the possibility of a simultaneous activations of a group of GCs (combined activity) in the retina by the output of amacrine cells [126–128], we suggest the possibility that the edge integration starts in the retina to some extent, in which it reflects in the retinal spike trains send to the cortex for the completion of the process.

We are not claiming that our simple multiple-scale DoG model can replace all the mid- to high-level explanations for similar illusions. We believe that many of these approaches have broad value, for example in contour illusion explanations or LBT illusions, which need further analysis or even previous knowledge and inferences for their explanation. What we have presented here builds on previous work [14, 48–50], highlighting the power of a first-stage multiple-scale model (with gradual changes of scale) based on retinal physiological insights and its multiscale processing of a visual scene. The aim was to show how much of visual information can be revealed and encoded by low-level retinal/cortical simple cells processing. This first-stage multiple-scale output can then feed to other high-level models for further application-based processing (for instance reweighting for normalization schemes or by the fusions of multiple scales into a multi-scale representation).

Here we have connected the output of our DoG model at multiple scales, to some prominent perceptual grouping principles such as continuity, connectivity and similarity, which are high-level perceptual explanations of our final percept. What we believe is that bioplausible low-level filtering techniques like 'Lateral Inhibitions' and 'Contrast Sensitivity' models are able to answer many Geometrical and Brightness/Lightness Illusions based on their nature of multiscale processing. We have shown here that simple circular centre and surround implementation of RGCs can reveal the emergence of tilt in the Tile Illusions. Although our model is a classical RF (CRF) implementation of RGC responses at multiple scales, the scales of the DoG edge map have been specified based on the characteristics of the investigated pattern. Additional implementations for orientation tuning of retinal ganglion cells (either nonlinear spatial subunits or nCRF implementations) might facilitate the investigations on broader range of Geometrical and Brightness/Lightness Illusions.

## 5 Conclusions and future work

We investigated the lateral interaction effect in visual processing of the perceptual organization of pattern elements and how it is connected to mid- and high-level result of grouping factors in our global percept, by simple computational modelling of retinal/cortical simple cells. In the perception of a visual scene, the grouping means 'putting items seen in the visual field together, or organizing image data' [129, p. 305]. It should be noted that the interest in grouping processes originated with the effort of Gestalt psychologists who have been the first to study grouping comprehensively as part of the general process of perception [130] (1920s, 1930s) and formulated a set of rules to explain the groupings perceived by humans. These grouping principles were typically justified by drawing parallels with certain neurological theories that were known at the time for their underlying neural mechanism.

In computational modelling of vision, Marr pointed out the need for perceptual clustering algorithms for obtaining full primal sketch from the raw primal sketch, using criteria such as collinearity and size similarity [22]. Clustering and segmentation algorithms broadly studied by the researchers in the field of CV and our intention for explaining illusory tilt effect focus on simple modelling of retinal low-level processing. We presented our investigation on a variant of the retinal classical receptive field (CRF) model implementing processing in the retinal ganglia and then used the model to generate an edge map representation as a raw primal sketch, which clearly highlight the emergence of tilt on a few Tile Illusion patterns.

It is well established that the centre–surround receptive fields of the RGCs are 'contrast estimators' [97]. We implemented our model based on a circular centre and surround organization of RGC responses considering the lateral inhibition interaction among them and utilized the DoG filtering at multiple scales to explain some of the second-order Tilt Illusions [14, 50] (referred to as Tile Illusions in our study), in particular 'Café Wall' and 'Complex Bulge' patterns here. We used this DoG edge map at multiple scales on Tile Illusion patterns described here for prediction and measurement of both tilt magnitude and its direction, by further analysis on their DoG edge maps by our model.

Although the edge map representation at multiple scales with gradual change of scale as the model output might seem over-complete, it has the potential to provide a lower error model of the data and is more likely to provide the information at the level of detail required for a particular image or application. One advantage of such models is that the model output is not sensitive to the exact parameter settings. It is important that the range of scales in the bioplausible DoG-based model is defined in such a way as to capture both high-frequency edge and texture details and the low-frequency profile information conveyed by brightness/colour from the objects within the scene. Thus, the neural processing of images that leads to the final potentially illusory output occurs at multiple scale levels. The number of such scales involved is a function of the parameters of the model. In general, reducing the number of scales is possible by increasing the incremental step, but this needs to be managed in a way that does not lose any intermediate or preliminary perceptual information that emerges during visual processing of a pattern or scene.

In this current work, we further investigated this model and the Tile Illusion patterns to find a connection of our 'bioderived Low-level filtering' model to some high-level 'Perceptual grouping' organizations with our main focus on Gestalt grouping principles, for example continuity (good continuation), connectivity and similarity. The experimental results show that the output of the model as an edge map representation of a stimuli could provide us not only the multiple-scale edge information as the indications for some shades around the edges, but also other hidden information such as local texture in the stimuli, as well as possible underlying mechanism for perceptual grouping arrangements at different scales. Therefore, the implementation of lateral inhibition in RGCs using DoG processing at multiple scales creates a feasible perceptual interpretation of the local structure in the pattern in a way that best meets its global perception.

We expect further that these low-level filtering approaches (retinal/cortical) have the ability to play a significant role in other higher-level models in relation to depth and motion processing which can contribute to the high-level top-down explanations of visual processing and can be extended to nCRF modelling for orientation selectivity of visual complex cells. Even the combination of the nCRF and the CRF for modelling RGCs could go a long way in supporting low-level filtering models. There is still much debate about the extent of coverage of isotropic filters in low-level approaches to represent the visual cues. Two successful models demonstrating the power of symmetrical filters in brightness perception models can be found in [131, 132]. Dakin and Bex [131] proposed a model for amplification of the low spatial frequency information of the image. Their model relies on a reconstruction phase based on the natural statistics of the image using a series of centre–surround, Laplacian of Gaussian filters. They demonstrated that complex models of orientation selective filters are not essential to successfully model brightness phenomenon such as White's effect and Craik–O'Brien–Cornsweet (COC) illusions. They highlighted the importance of normalization within low-level models rather than elongated filters. Zeman et al. [132] utilized a family of exponential filters (again isotropic) with multiple sizes and shapes instead of DoG/LoG for predicting the perceived brightness in their model. Their simulation results are comparable with the results achieved by the state-of-the-art Brightness/Lightness models [17, 67, 68] addressing both assimilation and contrast effects in Brightness/Lightness Illusions. All of these CV models certainly should get support from physiological evidence of retinal and cortical visual processing.

The extent of retinal encoding of visual input is not yet clear. The complexity of retinal processing of visual data is summarized by Field and Chichilnisky who note that: 'retinal spike trains are significantly more complex than was commonly appreciated, exhibiting surprisingly precise spike timing and highly structured concerted activity in different cells' [1, p. 2]. They also note that: 'three-pathway model fails to capture the functional diversity in the LGN. Furthermore, the specificity of connections from LGN to primary visual cortex extends beyond the well-known magnocellular/parvocellular separation [133–136]' [1, p. 11]. Important challenges of neurophysiological discoveries of the retinal encoding and circuitry are noted to be: 'identifying the distinct subcircuits that terminate on each RGC type, identifying the diversity of amacrine cell function and its contribution to shaping RGC responses, and identifying how RGCs within and across mosaics interact in communicating visual information to the brain' [1, p. 11].

The patterns we investigated are black and white, but individual cones are red, green or blue (M, L or S), and RGCs of different types exhibit different colour opponency properties and prevalences and thus different resolutions for different opponent colours. Thus, the contrasts for a black and white image that occur in the retina must be mediated by B–Y or R–G opponents, with the former being primary in mammals [137]. To explore the different spatial frequency characteristics of different RGC types, DeValois and DeValois [138] investigated both chromatic and achromatic versions of the checkerboard illusion. For chromatic and achromatic images *of the same size*, they demonstrated the assimilation in chromatic checkerboards where there was contrast for the achromatic version. They explained that assimilation effect is due to the chromatic system having much lower spatial frequencies compared to the achromatic system. For the Tile Illusions, Westheimer [139] investigate an isoluminant heterochromatic version of one sample Bulge pattern [109] and suggest that the illusion can disappear when the black and white tiles are replaced by isoluminant ones. Although this is not clear in the isoluminant version presented in the article, it can be expected for the above reasons at certain sizes and scales. More psychophysical experiments are needed to clarify the role of colour in Tile Illusions, including testing these illusions in both chromatic and achromatic versions on colour blind people as well as normal subjects. Shapiro [140] developed a quantitative model for the visual response of the cone cells with two separate pathways, one for luminance and chromatic information and one for contrast information. He presented the output of the model for disc-ring stimulus containing two discs of identical chromaticity and luminance, one surrounded by light ring and one surrounded by a dark ring. He showed that the contrast pathway appears to have a faster response compared to the colour pathway (Fig. 5 in [140]). A quantitative model of achromatic colour computation of similar stimulus used in [140] has been investigated by Rudd and Zemach [141] based on a distance-dependent edge integration mechanism. They have shown the outperformance of the edge integration model over the highest luminance rules of Anchoring theory [26].

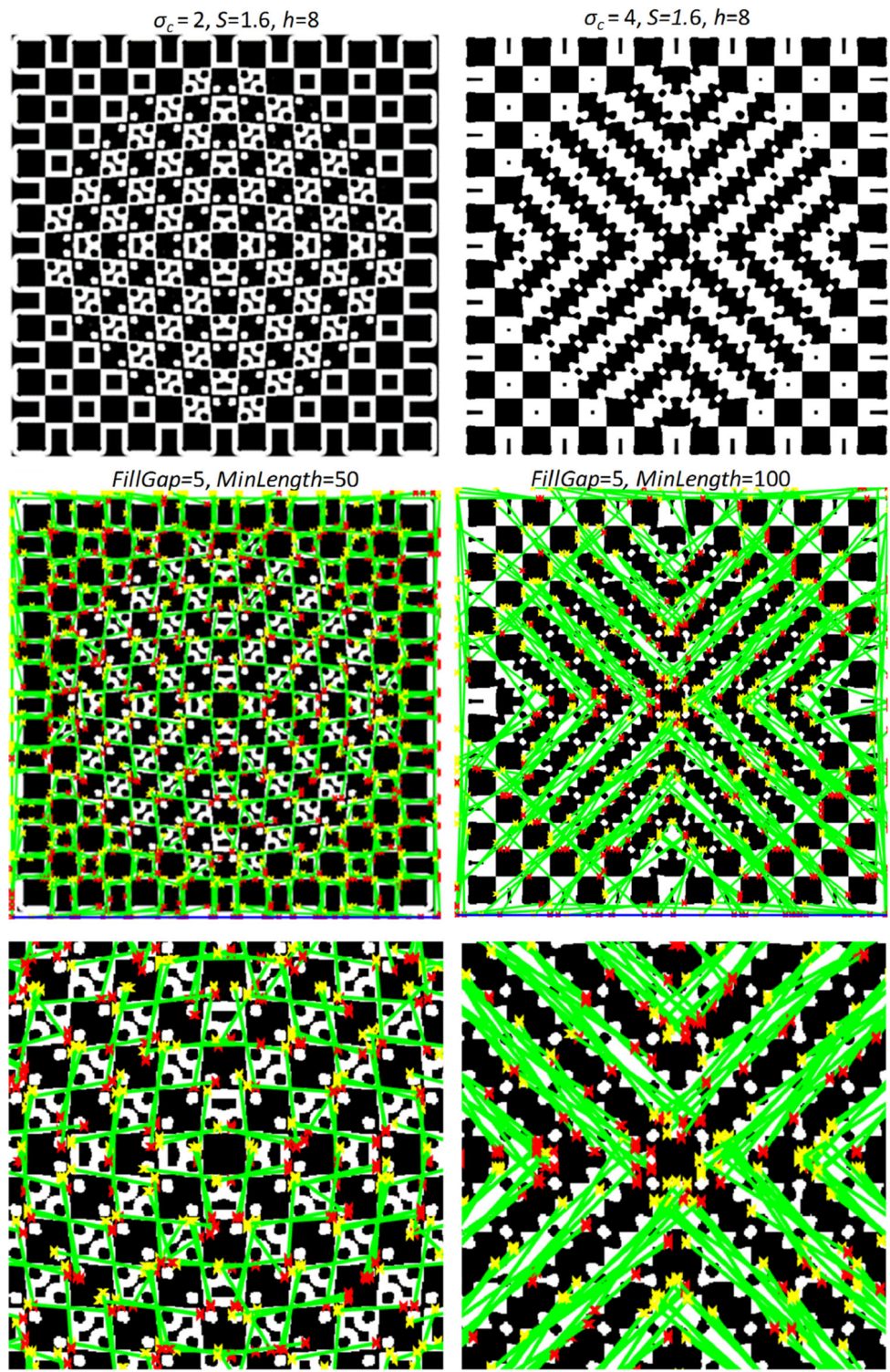

**Fig. 7** Detected houghlines (*centre*) for two scales of 2 and 4 ($\sigma_c = 2, 4$) from the edge map of the Complex Bulge pattern (*top*). The two images at the *bottom* of the figure are zoomed-in versions of the two images at the *centre row*. The parameters of the model and Hough investigation have been provided on the figure (Reproduced by permission from [145]). (Color figure online)

In conclusion, we have presented a detailed explanation of the extent to which DoG (and LoG) model predicts the Café Wall illusion specifically and measured the degree of tilt quantitatively in each scale utilizing Hough space [48–50]. We have also explored the concept of applying a second-stage processing model for orientation detection of tilt in broader range of Tilt Illusions such as Bulge patterns. The measurement of tilt value should consider both global tilt measurement (overall view of the pattern) and the 'local focus on tilt' predicting local tilt percept or the edge displacement. This will highlight a more precise connection of our Tilt Illusion explanation to the perceptual grouping of visual data. We need to estimate the illusion strength and orientation based on a psychophysical assessment of the model prediction, which is a priority in our future study.

In our future work, we intend to add orientation selectivity to the model and make an extension to an implementation of a nonlinear spatial subunits and/or a nCRFs model, inspired by biological findings [95, 142–144]. These extensions can be achieved either by using a summation of Gaussian components at multiple spatial scales, or nCRF implementations. This may be achieved by isotropic filters with three Gaussians such as extended CRF model of Ghosh et al., with classical excitatory and inhibitory surround Gaussians, and a non-classical extended disinhibitory field (surround) [65, 99], or by anisotropic filters with elongated surrounds in different orientations such as the Brightness/Lightness model of Blakeslee and McCourt [31, 67]. The aim will be to explore more on orientation tuning cells and their effects on the perceived brightness, with an attempt to connect the edge map of our model to brightness models, and most importantly to design a bioderived second-stage processing in our model for identifying angles of orientation on detected tilts in the edge map quantitatively instead of the current Hough analysis stage. This extension to the current model facilitates further processing of the revealed tilt in the edge map representation of the patterns. Our intention would be to build our analytical model similar to our visual processing for searching of different visual clues in natural or illusion patterns.

**Acknowledgements** Nasim Nematzadeh was supported by an Australian Research Training Program (RTP) Award. The authors would like to appreciate the precise and useful comments of the reviewers on the first draft of our manuscript which lead to goal-directed improvement in the final version of our presented research. The authors assert that they have no conflict of interest in relation to the research reported.

# Appendix

See Tables 1 and 2.

**Table 1** Geometrical and Brightness/Lightness Illusion patterns (Reproduced by permission from [145]). The source and original references of illusions in this table are provided in Table 2

| | a | b | c | d |
|---|---|---|---|---|
| 1 | Mach Bands | Chevreul | Pyramid | Grating Induction |
| 2 | Craik-O'Brien-Cornsweet (COC) | Irradiation | Simultaneous Brightness Contrast (SBC) | White's effect |
| 3 | Hermann Grid | Café Wall | Fraser | Twisted Cord |
| 4 | Spring | Spiral Café Wall | Spiral Fraser | Sound Wave |
| 5 | Zöllner | Herring-Wundt | Orbison | Poggendorff |
| 6 | Ebbinghaus | Müller-Lyer | Ponzo | Kanizsa Square |
| 7 | Face - Vase | Necker Cube | Penrose Triangle | Impossible Staircase |
| 8 | Knill & Kersten | Wall of Block | Crisscross | Snake |

Table 2 The source and original references of the illusion patterns in Table 1

| Illusion patterns | Appendix Row#- a…d | Source/original Author (date) | References Source/original |
|---|---|---|---|
| Mach Bands | R1-a | Penacchio et al. (2013)/Mach (1865), Fiorentini (1972) | [63]/[146, 147] |
| Chevruell | R1-b | Geier (2011)/Chevreul (1890) | [148]/[149] |
| Pyramid | R1-c | Troncoso et al. (2005)/Vasarely (1966, 1970) | [150]/[151, 152] |
| Grating Induction (GI) | R1-d | Penacchio et al. (2013)/McCourt (1982) | [63]/[153] |
| Craik–O'Brien Cornsweet (COC) | R2-a | Web Img, Lu & Sperling (1995)/Craik (1940), O'Brien (1958), Cornsweet (1970) | [154, 155]/[156–158] |
| Irradiation | R2-b | Westheimer (2007)/Helmholtz (1896) | [45]/[62] |
| Simultaneous Brightness Contrast (SBC) | R2-c | Adelson (2000)/Heinemann (1955) | [19]/[159] |
| White's effect (WE) | R2-d | Blakeslee and McCourt (1999)/White (1979) | [17]/[15] |
| Herman Grid | R3-a | Blakeslee and McCourt (1997)/Hermann (1870) | [106]/[160] |
| Café Wall | R3-b | Nematzadeh (2015)/Gregory (1973), Munsterberg (1897), Pierce (1898), Fraser (1908) | [50]/[161–163, 33] |
| Fraser | R3-c | Kitaoka (2007)/Fraser (1908) | [71]/[33] |
| Twisted cord | R3-d | McCourt (1983)/Fraser (1908) | [41]/[33] |
| Spring | R4-a | Fermuller & Malm (2004)/Kitaoka (2003) | [107]/[164] |
| Spiral Café Wall | R4-b | Kitaoka (2007)/Kitaoka et al. (2001) | [71]/[165] |
| Spiral Fraser | R4-c | Kitaoka (2007)/Gregory & Heard (1979), Kitaoka et al. (2001) | [71]/[42, 165] |
| Sound Wave | R4-d | Stevanov et al. (2012)/Kitaoka (1998) | [166]/[109] |
| Zöllner | R5-a | Web–Wikipedia/Zöllner (1862) | [167]/[168] |
| Herring–Wundt | R5-b | Changizi et al. (2008)/Herring (1861), Wundt (1898) | [9]/[169, 170] |
| Orbison | R5-c | Changizi et al. (2008)/Orbison (1939) | [9]/[171] |
| Poggendorff | R5-d | Ninio (2014)/Zöllner (1860) | [8]/[172] |
| Ebbinghaus | R6-a | Ninio (2014)/Ebinghaus (1902) | [8]/[173] |
| Müller–Lyer | R6-b | Ninio (2014)/Müller-Lyer (1889) | [8]/[174] |
| Ponzo | R6-c | Ninio (2014)/Ponzo (1910) | [8]/[175] |
| Kanizsa Triangle | R6-d | Eagleman (2001)/Kanizsa (1974), Frisby & Clatworthy (1975) | [13]/[176, 177] |
| Face–Vase | R7-a | Sturmberg (2011)/Rubin (1915) | [178]/[179] |
| Necker Cube | R7-b | Sturmberg (2011)/Necker (1832) | [178]/[180] |
| Penrose Triangle | R7-c | Web—OpenClipArt/Pappas (1989) | [181]/[182] |
| Impossible Staircase | R7-d | Web—Wikipedia/Penrose (1958) | [183]/[184] |
| Knill & Kersten | R8-a | Adelson (2000)/Knill & Kresten (1991) | [19]/[185] |
| Wall of Block | R8-b | Logvinenko et al. (2002)/Adelson (1993) | [21]/[186] |
| Crisscross | R8-c | Adelson (2000) | [19] |
| Snake | R8-d | Adelson (2000) | [19] |

## References

1. Field G, Chichilnisky E (2007) Information processing in the primate retina: circuitry and coding. Annu Rev Neurosci 30:1–30
2. Gollisch T, Meister M (2010) Eye smarter than scientists believed: neural computations in circuits of the retina. Neuron 65(2):150–164
3. Draper SW (1978) The Penrose triangle and a family of related figures. Perception 7(3):283–296
4. Wright AS (2013) The origins of Penrose diagrams in physics, art, and the psychology of perception, 1958–62. Endeavour 37(3):133–139
5. Cowan TM (1982) Turning a Penrose triangle inside out. J Math Psychol 26(3):252–262
6. Kornmeier J, Bach M (2005) The Necker cube—an ambiguous figure disambiguated in early visual processing. Vis Res 45(8):955–960
7. Prinzmetal W, Beck DM (2001) The tilt-consistency theory of visual illusions. J Exp Psychol Hum Percept Perform 27(1):206
8. Ninio J (2014) Geometrical illusions are not always where you think they are: a review of some classical and less classical illusions, and ways to describe them. Front Hum Neurosci 8. doi:10.3389/fnhum.2014.00856
9. Changizi MA, Hsieh A, Nijhawan R, Kanai R, Shimojo S (2008) Perceiving the present and a systematization of illusions. Cognit Sci 32(3):459–503

10. Ratliff F (1965) Mach bands: quantitative studies on neural networks. Holden-Day, San Francisco
11. Yantis S (2013) Sensation and perception. Palgrave Macmillan, Basingstoke
12. Gregory RL (1997) Knowledge in perception and illusion. Philos Trans R Soc B Biol Sci 352(1358):1121–1127
13. Eagleman DM (2001) Visual illusions and neurobiology. Nat Rev Neurosci 2(12):920–926
14. Nematzadeh N, Lewis TW, Powers DM (2015) Bioplausible multiscale filtering in retinal to cortical processing as a model of computer vision. In: ICAART2015-international conference on agents and artificial intelligence. Scitepress, Lisbon
15. White M (1979) A new effect of pattern on perceived lightness. Perception 8(4):413–416
16. Howe P (2001) Explanations of White's effect that are based solely on T-junctions are incomplete. In: Investigative ophthalmology & visual science. Assoc research vision ophthalmology Inc 9650 Rockville pike, Bethesda, MD 20814-3998, USA
17. Blakeslee B, McCourt ME (1999) A multiscale spatial filtering account of the White effect, simultaneous brightness contrast and grating induction. Vis Res 39(26):4361–4377
18. Wallach H (1963) The perception of neutral colors. Scientific American, Armonk
19. Adelson EH (2000) Lightness perception and lightness illusions. In: Gazzaniga (ed) The new cognitive neurosciences, 2nd edn. MIT Press, Cambridge, MA, pp 339–351
20. Logvinenko AD, Kane J (2004) Hering's and Helmholtz's types of simultaneous lightness contrast. J Vis 4(12):9
21. Logvinenko AD, Kane J, Ross DA (2002) Is lightness induction a pictorial illusion? Perception 31(1):73–82
22. Marr D (1982) Vision: a computational investigation into the human representation and processing of visual information. W.H. Freeman, New York
23. Grossberg S, Todorovic D (1988) Neural dynamics of 1-D and 2-D brightness perception: a unified model of classical and recent phenomena. Percept Psychophys 43(3):241–277
24. Ross WD, Pessoa L (2000) Lightness from contrast: a selective integration model. Percept Psychophys 62(6):1160–1181
25. Kellman P (2003) Interpolation processes in the visual perception of objects. Neural Netw 16(5):915–923
26. Gilchrist A, Kossyfidis C, Bonato F, Agostini T, Cataliotti J, Li X, Spehar B, Annan V, Economou E (1999) An anchoring theory of lightness perception. Psychol Rev 106(4):795
27. Todorovic D (1997) Lightness and junctions. Perception 26:379–394
28. Anderson BL (1997) A theory of illusory lightness and transparency in monocular and binocular images: the role of contour junctions. Perception 26(4):419–454
29. Anderson BL, Winawer J (2005) Image segmentation and lightness perception. Nature 434(7029):79–83
30. Kingdom F, Moulden B (1992) A multi-channel approach to brightness coding. Vis Res 32(8):1565–1582
31. Blakeslee B, McCourt ME (2003) A multiscale spatial filtering account of brightness phenomena. In: Harris L, Jenkin M (eds) Levels of perception. Springer, New York, pp 47–72
32. Kingdom FA (2011) Lightness, brightness and transparency: a quarter century of new ideas, captivating demonstrations and unrelenting controversy. Vis Res 51(7):652–673
33. Fraser J (1908) A new visual illusion of direction. Br J Psychol 2(3):307–320
34. Lennie P (1981) The physiological basis of variations in visual latency. Vis Res 21(6):815–824
35. Maunsell JH, Gibson JR (1992) Visual response latencies in striate cortex of the macaque monkey. J Neurophysiol 68(4):1332–1344
36. Ramachandran VS, Anstis SM (1990) Illusory displacement of equiluminous kinetic edges. Perception 19(5):611–616
37. Nijhawan R (1994) Motion extrapolation in catching. Nature 370:256–257
38. Nijhawan R (2002) Neural delays, visual motion and the flash-lag effect. Trends Cognit Sci 6(9):387–393
39. Changizi MA, Widders DM (2002) Latency correction explains the classical geometrical illusions. Perception 31(10):1241–1262
40. Briscoe RE (2010) Perceiving the present: systematization of illusions or illusion of systematization? Cognit Sci 34(8):1530–1542
41. McCourt ME (1983) Brightness induction and the Café Wall illusion. Perception 12(2):131–142
42. Gregory RL, Heard P (1979) Border locking and the Café Wall illusion. Perception 8(4):365–380
43. Tani Y, Maruya K, Sato T (2006) Reversed Café Wall illusion with missing fundamental gratings. Vis Res 46(22):3782–3785
44. Earle DC, Maskell SJ (1993) Fraser cords and reversal of the Café Wall illusion. Perception 22:383–390
45. Westheimer G (2007) Irradiation, border location, and the shifted-chessboard pattern. Perception 36(4):483
46. Morgan M, Moulden B (1986) The Münsterberg figure and twisted cords. Vis Res 26(11):1793–1800
47. Moulden B, Renshaw J (1979) The Munsterberg illusion and 'irradiation'. Perception 8:275–301. doi:10.1068/p08027
48. Nematzadeh N, Powers DM (2016) A quantitative analysis of tilt in the Café Wall illusion: a bioplausible model for foveal and peripheral vision. In: Digital image computing: techniques and applications (DICTA). IEEE, pp 1–8
49. Nematzadeh N, Powers DM, Trent L (2016) Quantitative analysis of a bioplausible model of misperception of slope in the Café Wall illusion. In: Workshop on interpretation and visualization of deep neural nets (WINVIZNN) ACCV
50. Nematzadeh N, Powers DM (2016) A bioplausible model for explaining Café Wall illusion: foveal versus peripheral resolution. In: International symposium on visual computing. Springer, Berlin, pp 426–438
51. Jameson D (1985) Opponent-colours theory in the light of physiological findings. In: Central and peripheral mechanisms of colour vision. Palgrave Macmillan, UK, pp 83–102
52. Smith VC, Jin PQ, Pokorny J (2001) The role of spatial frequency in color induction. Vis Res 41(8):1007–1021
53. Arai H (2005) A nonlinear model of visual information processing based on discrete maximal overlap wavelets. Interdiscip Inf Sci 11(2):177–190
54. Dixon E, Shapiro A, Lu ZL (2014) Scale-invariance in brightness illusions implicates object-level visual processing. Sci Rep 4:3900. doi:10.1038/srep03900
55. Burt PJ, Adelson EH (1983) The Laplacian pyramid as a compact image code. IEEE Trans Commun 31(4):532–540
56. Mallat S (1996) Wavelets for a vision. Proc IEEE 84(4):604–614
57. Lowe DG (1999) Object recognition from local scale-invariant features. In: The proceedings of the seventh IEEE international conference on computer vision, vol 2. IEEE, pp 1150–1157
58. Lindeberg T (2011) Generalized Gaussian scale-space axiomatics comprising linear scale-space, affine scale-space and spatio-temporal scale-space. J Math Imaging Vis 40(1):36–81
59. Jacques L, Duval L, Chaux C, Peyré G (2011) A panorama on multiscale geometric representations, intertwining spatial, directional and frequency selectivity. Signal Process 91(12):2699–2730
60. Lourens T (1995) Modeling retinal high and low contrast sensitivity filters. In: From natural to artificial neural computation. Springer, Berlin, pp 61–68
61. Romeny BMH (2003) Front-end vision and multi-scale image analysis: multi-scale computer vision theory and applications, written in Mathematica, vol 27. Springer, Berlin

62. von Helmholtz H (1911) Handbuch der Physiologischen, Optik, vol II. In: Southall JPC (ed) Helmoholc's Treatise on Physiological. Optics, 1962, vols I, II. Dover, New York
63. Penacchio O, Otazu X, Dempere-Marco L (2013) A neurodynamical model of brightness induction in V1. PLoS ONE 8(5):e64086
64. Shapiro A, Lu ZL (2011) Relative brightness in natural images can be accounted for by removing blurry content. Psychol Sci 22:1452–1459
65. Ghosh K, Sarkar S, Bhaumik K (2006) A possible explanation of the low-level brightness–contrast illusions in the light of an extended classical receptive field model of retinal ganglion cells. Biol Cybern 94(2):89–96
66. Moulden B, Kingdom F (1989) White's effect: a dual mechanism. Vis Res 29(9):1245–1259
67. Blakeslee B, McCourt ME (2004) A unified theory of brightness contrast and assimilation incorporating oriented multiscale spatial filtering and contrast normalization. Vis Res 44(21):2483–2503
68. Robinson AE, Hammon PS, de Sa VR (2007) Explaining brightness illusions using spatial filtering and local response normalization. Vis Res 47(12):1631–1644
69. Yu, Y., Yamauchi T., Choe Y (2004 Explaining low-level brightness-contrast illusions using disinhibition. In: International workshop on biologically inspired approaches to advanced information technology. Springer, Berlin, pp 166–175
70. Kitaoka A (2000) Trampoline pattern (web image). http://www.ritsumei.ac.jp/~akitaoka/motion-e.html
71. Kitaoka A (2007) Tilt illusions after Oyama (1960): a review. Jpn Psychol Res 49(1):7–19
72. Illingworth J, Kittler J (1988) A survey of the Hough transform. Comput Vis Graph Image Process 44(1):87–116
73. Marr D, Hildreth E (1980) Theory of edge detection. Proc R Soc Lond Ser B Biol Sci 207(1167):187–217
74. Shapley R, Perry VH (1986) Cat and monkey retinal ganglion cells and their visual functional roles. Trends Neurosci 9:229–235
75. Martinez-Conde S, Macknik SL, Hubel DH (2004) The role of fixational eye movements in visual perception. Nat Rev Neurosci 5(3):229–240
76. Bressan P (2006) The place of white in a world of grays: a double-anchoring theory of lightness perception. Psychol Rev 113(3):526
77. Wagemans J, Elder JH, Kubovy M, Palmer SE, Peterson MA, Singh M, von der Heydt R (2012) A century of gestalt psychology in visual perception: I. Perceptual grouping and figure–ground organization. Psychol Bull 138(6):1172
78. Wagemans J, Feldman J, Gepshtein S, Kimchi R, Pomerantz JR, van der Helm PA, van Leeuwen C (2012) A century of Gestalt psychology in visual perception: II. Conceptual and theoretical foundations. Psychol Bull 138(6):1218
79. Spillmann L, Werner JS (eds) (2012) Visual perception: the neurophysiological foundations. Elsevier, Amsterdam
80. Bruce V, Green P, Georgeson M (1996) Visual perception: physiology, psychology, and ecology. Lawrence Earlbaum Associates, Hove
81. Carlson NR, Buskist W, Enzle ME, Heth CD (2000) Psychology: the science of behaviour. Allyn and Bacon, Scarborough
82. Wagemans J (2014) How much of Gestalt theory has survived a century of neuroscience. Perception beyond gestalt: progress in vision research. Psychology Press, New York, NY, pp 9–21
83. Hochstein S, Ahissar M (2002) View from the top: hierarchies and reverse hierarchies in the visual system. Neuron 36(5):791–804
84. Bar M, Kassam KS, Ghuman AS, Boshyan J, Schmid AM, Dale AM, Halgren E (2006) Top-down facilitation of visual recognition. Proc Natl Acad Sci USA 103(2):449–454
85. Spillmann L, Dresp-Langley B, Tseng CH (2015) Beyond the classical receptive field: the effect of contextual stimuli. J Vis 15(9):7
86. Grossberg S, Mingolla E (1985) Neural dynamics of perceptual grouping: textures, boundaries, and emergent segmentations. Attent Percept Psychophys 38(2):141–171
87. Spillmann L, Werner JS (1996) Long-range interactions in visual perception. Trends Neurosci 19(10):428–434
88. Field DJ, Hayes A, Hess RF (1993) Contour integration by the human visual system: evidence for a local association field. Vis Res 33(2):173–193
89. Gilbert CD, Li W (2013) Top-down influences on visual processing. Nat Rev Neurosci 14(5):350–363
90. Craft E, Schütze H, Niebur E, Von Der Heydt R (2007) A neural model of figure–ground organization. J Neurophysiol 97(6):4310–4326
91. Roelfsema PR (2006) Cortical algorithms for perceptual grouping. Annu Rev Neurosci 29:203–227
92. Hubel DH, Wiesel TN (1962) Receptive fields, binocular interaction and functional architecture in the cat's visual cortex. J Physiol 160(1):106–154
93. Barlow HB (1953) Summation and inhibition in the frog's retina. J Physiol 119(1):69–88
94. Kuffler SW (1952) Neurons in the retina: organization, inhibition and excitation problems. In: Cold Spring Harbor symposia on quantitative biology. Cold Spring Harbor Laboratory Press, New York
95. Passaglia CL, Enroth-Cugell C, Troy JB (2001) Effects of remote stimulation on the mean firing rate of cat retinal ganglion cells. J Neurosci 21(15):5794–5803
96. Rodieck RW, Stone J (1965) Analysis of receptive fields of cat retinal ganglion cells. J Neurophysiol 28(5):833–849
97. Enroth-Cugell C, Robson JG (1966) The contrast sensitivity of retinal ganglion cells of the cat. J Physiol 187(3):517–552
98. Pessoa L (1996) Mach bands: how many models are possible? Recent experimental findings and modeling attempts. Vis Res 36(19):3205–3227
99. Ghosh K, Sarkar S, Bhaumik K (2009) A possible mechanism of stochastic resonance in the light of an extra-classical receptive field model of retinal ganglion cells. Biol Cybern 100(5):351–359
100. Powers DMW (1983) lateral interaction behaviour derived from neural packing considerations. University of New South Wales, School of Electrical Engineering and Computer Science, Sydney
101. Malsburg C (1973) Self-organization of orientation sensitive cells in the striate cortex. Kybernetik 14(2):85–100
102. Economou E, Zdravkovic S, Gilchrist A (2007) Anchoring versus spatial filtering accounts of simultaneous lightness contrast. J Vis 7(12):2
103. Otazu X, Vanrell M, Párraga CA (2008) Multiresolution wavelet framework models brightness induction effects. Vis Res 48(5):733–751
104. Gilchrist AL, Radonjić A (2010) Functional frameworks of illumination revealed by probe disk technique. J Vis 10(5):6
105. Ghosh K, Sarkar S, Bhaumik K (2007) The theory of edge detection and low-level vision in retrospect. INTECH Open Access Publisher, New York
106. Blakeslee B, McCourt ME (1997) Similar mechanisms underlie simultaneous brightness contrast and grating induction. Vis Res 37(20):2849–2869
107. Fermüller C, Malm H (2004) Uncertainty in visual processes predicts geometrical optical illusions. Vis Res 44(7):727–749
108. Kitaoka A, Pinna B, Brelstaff G (2004) Contrast polarities determine the direction of Café Wall tilts. Perception 33(1):11–20
109. Kitaoka A (1998) Apparent contraction of edge angles. Perception 27(10):1209–1219

110. Lulich DP, Stevens KA (1989) Differential contributions of circular and elongated spatial filters to the Café Wall illusion. Biol Cybern 61(6):427–435
111. Mangel SC (1991) Analysis of the horizontal cell contribution to the receptive field surround of ganglion cells in the rabbit retina. J Physiol 442(1):211–234
112. Linsenmeier RA, Frishman LJ, Jakiela HG, Enroth-Cugell C (1982) Receptive field properties of X and Y cells in the cat retina derived from contrast sensitivity measurements. Vis Res 22(9):1173–1183
113. Young R (1985) The Gaussian derivative theory of spatial vision: analysis of cortical cell receptive field line-weighting profiles. Publication GMR-4920, General Motors Research Labs. Computer Science Department 30500, pp 48090–49055
114. Young RA (1987) The Gaussian derivative model for spatial vision: I. Retinal mechanisms. Spat Vis 2(4):273–293
115. Ghosh K, Sarkar S, Bhaumik K (2007) Understanding image structure from a new multi-scale representation of higher order derivative filters. Image Vis Comput 25(8):1228–1238
116. Lindeberg T, Florack L (1994) Foveal scale-space and the linear increase of receptive field size as a function of eccentricity. Technical report. ISRN: KTH NA/P-94/27-SE
117. Bay H, Ess A, Tuytelaars T, Van Gool L (2008) Speeded-up robust features (SURF). Comput Vis Image Underst 110(3):346–359
118. Hamid N, Yahya A, Ahmad RB, Al-Qershi O (2012) Characteristic region based image steganography using speeded-up robust features technique. In: 2012 international conference on future communication networks (ICFCN). IEEE
119. Xu Y, Huang S, Ji H, Fermüller C (2012) Scale-space texture description on SIFT-like textons. Comput Vis Image Underst 116(9):999–1013
120. Smith S (2013) Digital signal processing: a practical guide for engineers and scientists. Newnes, Oxford
121. Grossberg S, Mingolla E (1985) Neural dynamics of form perception: boundary completion, illusory figures, and neon color spreading. Psychol Rev 92(2):173
122. Nematzadeh N, Powers DM (2017) Modeling geometrical mysteries of Café Wall illusions. Retitled as: A productive account of Café Wall illusions using quantitative model. https://arxiv.org/abs/1705.06846 (in press)
123. Barlow HB, Hill RM (1963) Selective sensitivity to direction of movement in ganglion cells of the rabbit retina. Science 139:412–414
124. Barlow HB, Hill RM (1963) Evidence for a physiological explanation of the waterfall phenomenon and figural after-effects. Nature 200(4913):1345–1347
125. Weng S, Sun W, He S (2005) Identification of ON–OFF direction-selective ganglion cells in the mouse retina. J Physiol 562(3):915–923
126. Barlow HB, Derrington AM, Harris LR, Lennie P (1977) The effects of remote retinal stimulation on the responses of cat retinal ganglion cells. J Physiol 269:177–194
127. Frishman LJ, Linsenmeier RA (1982) Effects of picrotoxin and strychnine on nonlinear responses of Y-type cat retinal ganglion cells. J Physiol 324:347–363
128. Roska B, Werblin F (2003) Rapid global shifts in natural scenes block spiking in specific ganglion cell types. Nat Neurosci 6:600–608
129. Ahuja N, Tuceryan M (1989) Extraction of early perceptual structure in dot patterns: integrating region, boundary, and component gestalt. Comput Vis Graph Image Process 48(3):304–356
130. Wertheimer M (1938) Laws of organization in perceptual forms. In: Ellis WD (ed) A source book of gestalt psychology. Harcourt Brace, New York, pp 71–88
131. Dakin SC, Bex PJ (2003) Natural image statistics mediate brightness 'filling in'. Proc R Soc Lond B Biol Sci 270(1531):2341–2348
132. Zeman A, Brooks KR, Ghebreab S (2015) An exponential filter model predicts lightness illusions. Front Hum Neurosci 9:368. doi:10.3389/fnhum.2015.00368
133. Chatterjee S, Callaway EM (2003) Parallel colour-opponent pathways to primary visual cortex. Nature 426(6967):668–671
134. Sun H, Ruttiger L, Lee BB (2004) The spatiotemporal precision of ganglion cell signals: a comparison of physiological and psychophysical performance with moving gratings. Vis Res 44:19–33
135. Callaway EM (2005) Structure and function of parallel pathways in the primate early visual system. J Physiol 566:13–19
136. Hendry SH, Reid RC (2000) The koniocellular pathway in primate vision. Annu Rev Neurosci 23:127–153
137. Martin PR (1998) Colour processing in the primate retina: recent progress. J Physiol 513(3):631–638
138. De Valois RL, De Valois KK (1993) A multi-stage color model. Vis Res 33(8):1053–1065
139. Westheimer G (2008) Illusions in the spatial sense of the eye: geometrical–optical illusions and the neural representation of space. Vis Res 48(20):2128–2142
140. Shapiro AG (2008) Separating color from color contrast. J Vis 8(1):8
141. Rudd ME, Zemach IK (2005) The highest luminance anchoring rule in achromatic color perception: some counterexamples and an alternative theory. J Vis 5(11):5
142. Carandini M (2004) Receptive fields and suppressive fields in the early visual system. Cognit Neurosci 3:313–326
143. Cavanaugh JR, Bair W, Movshon JA (2002) Nature and interaction of signals from the receptive field center and surround in macaque V1 neurons. J Neurophysiol 88(5):2530–2546
144. Tanaka H, Ohzawa I (2009) Surround suppression of V1 neurons mediates orientation-based representation of high-order visual features. J Neurophysiol 101(3):1444–1462
145. Nematzadeh N, A Neurophysiological model for geometric visual illusions. PhD Thesis, Flinders University (in preparation)
146. Mach E (1865) Ueber die physiologische Wirkung räumlichen Vertheilung des Lichtreize auf die Netzhaut. Sitzungsberichte der mathematisch-naturwissenschaftlichen Classe der kaiserlichen Akademie der Wissenschaften 52:303–322
147. Fiorentini A (1972) Mach band phenomena. In: Jameson D, Hurvich LM (eds) Handbook of sensory physiology, vol VII/4. Springer, Berlin, pp 188–201
148. Geier J, Hudák M (2011) Changing the Chevreul illusion by a background luminance ramp: lateral inhibition fails at its traditional stronghold-a psychophysical refutation. PLoS ONE 6(10):e26062
149. Chevreul ME (1967) Principles of harmony and contrast of colours (1839) reprinted, Introduction and noted by F. Birren. Van Nostrand Reinhold, New York
150. Troncoso XG, Macknik SL, Martinez-Conde S (2005) Novel visual illusions related to Vasarely's 'nested squares' show that corner salience varies with corner angle. Perception 34(4):409–420
151. Vasarely V (1966) Vasarely: Hayden Gallery, Massachusetts Institute of Technology, February 14 through March 20, 1966. MIT, Cambridge
152. Vasarely V, Joray M (1970) Vasarely II. Editions du Griffon, Neuchâtel
153. McCourt ME (1982) A spatial frequency dependent grating-induction effect. Vis Res 22(1):19–134
154. http://hboyaci.bilkent.edu.tr/Vision/
155. Lu ZL, Sperling G (1996) Second-order illusions: mach bands, chevreul, and Craik–O'Brien–Cornsweet. Vis Res 36(4):559–572

156. Craik KJW (1940) Visual adaptation. Doctoral dissertation, University of Cambridge
157. O'Brien V (1958) Contour perception, illusion and reality. JOSA 48(2):112–119
158. Cornsweet TN (1970) Vision perception. Academic Press, New York
159. Heinemann EG (1955) Simultaneous brightness induction as a function of inducing and test-field luminances. J Exp Psychol 50:8996
160. Hermann L (1870) Eine erscheinung simultanen contrastes. Pflügers Arch Eur J Physiol 3(1):13–15
161. Gregory RL (1973) The confounded eye. In: Gregory RL, Gombrich EH (eds) Illusion in nature and art. Duckworth, London, pp 49–95
162. Munsterberg H (1897) Die verschobene Schachbrettfigur. Z Psychol 15:184–188
163. Pierce AH (1898) The illusion of the kindergarten patterns. Psychol Rev 5(3):233
164. Kitaoka A (2003) Phenomenal characteristics of the peripheral drift illusion. Vision 15:261–262
165. Kitaoka A, Pinna B, Brelstaff G (2001) New variations of the spiral illusion. Perception 30(5):637–646
166. Stevanov J, Marković S, Kitaoka A (2012) Aesthetic valence of visual illusions. i-Perception 3(2):112–140
167. https://upload.wikimedia.org/wikipedia/commons/2/2d/Zollner_illusion.svg
168. Zöllner F (1862) Über eine neue Art anorthoskopischer Zerrbilder. Ann Phys 193(11):477–484
169. Hering E (1861) Der ortssinn der netzhaut. Engelmann [JTE], Lemgo
170. Wundt WM (1898) Die geometrisch-optischen Täuschungen, vol 24, 2nd edn. BG Teubner, Stuttgart
171. Orbison WD (1939) Shape as a function of the vector-field. Am J Psychol 52(1):31–45
172. Zöllner F (1860) Ueber eine neue Art von Pseudoskopie und ihre Beziehungen zu den von Plateau und Oppel beschriebenen Bewegungsphänomenen. Ann Phys 186(7):500–523
173. Ebbinghaus H (1902) The principles of psychology. Veit, Leipzig
174. Müller-Lyer FC (1889) Optische urteilstäuschungen. Arch Anat Phys Physiol Abt 2(Supplement):263–270
175. Ponzo M (1910) Intorno ad alcune illusioni nel campo delle sensazioni tattili, sull'illusione di Aristotele e fenomeni analoghi. Wilhelm Engelmann, Lemgo
176. Kanizsa G (1974) Contours without gradients or cognitive contours?. Ital J Psychol 1:93–113
177. Frisby JP, Clatworthy JL (1975) Illusory contours: curious cases of simultaneous brightness contrast? Perception 4(3):349–357
178. Sturmberg JP (2011) The illusion of certainty—a deluded perception? J Eval Clin Pract 17(3):507–510
179. Rubin E (1915) Synsoplevede figurer (visually experienced figures). Gyldendal, Copenhagen
180. Necker, LA (1832) LXI. Observations on some remarkable optical phænomena seen in Switzerland; and on an optical phænomenon which occurs on viewing a figure of a crystal or geometrical solid. Philos Mag Ser 3 1(5):329–337
181. https://openclipart.org/download/204324/3D-Triangle-illusion.svg
182. Pappas T (1989) Napoleon's theorem, the joy of mathematics. Wide World Publ./Tetra, San Carlos
183. https://upload.wikimedia.org/wikipedia/commons/3/34/Impossible_staircase.svg
184. Penrose LS, Penrose R (1958) Impossible objects: a special type of visual illusion. Br J Psychol 49(1):31–33
185. Knill DC, Kersten D (1991) Apparent surface curvature affects lightness perception. Nature 351(6323):228
186. Adelson EH (1993) Perceptual organization and the judgment of brightness. Science 262(5142):2042–2044

**Nasim Nematzadeh** undertook a Bachelor of Software Engineering at Amirkabir University of Technology (AUT), Tehran, Iran and completed her Masters of Computer Engineering in Artificial Intelligence and Robotics at the Science and Research University, Tehran, Iran. She is currently a PhD student in the college of Science and Engineering, Flinders University of South Australia. Her research interests are focused on bioplausible vision models, computer vision techniques, image processing and machine learning for geometrical illusions, as well as more broadly in applications of computational intelligence for visual data representation and modelling low-level vision.

**David M. W. Powers** is Professor of Computer and Cognitive Science at Flinders University in South Australia. He has research interests in the area of Artificial Intelligence and Cognitive Science. His specific research framework takes Language, Logic and Learning as the cornerstones for a broad Cognitive Science perspective on Artificial Intelligence and its practical applications. Prof. Powers is known as a pioneer in the area of Natural Language Learning, Unsupervised Learning and Evaluation of Learning, and was Founding President of ACLSIGNLL, as well as initiating the CoNLL conference. His previous positions include Telecom Paris, University of Tilburg, University of Kaiserslautern, Macquarie University, as well as working in industry, and commercialization of research through several start up companies.

**Trent W. Lewis** completed a Bachelor of Science with a major in Cognitive Science and Honours in Computer Science. This was followed by PhD examining audio-visual speech processing, predominately by machines, but also investigating how humans perceive and integrate speech from different modalities to create more elegant computational algorithms including research in dynamic weighting, classifier combination, confidence estimation and combination, and the use of linguistic distinctive features as units for fusion of audio visual speech. He worked as a post-doctoral researcher on the Thinking Head project – an ARC/NH&MRC Special Research Initiative that targeted the development of intelligent Embodied Conversational Agents. He is currently a Lecturer at Flinders University and part of the College of Science and Engineering, Brain Signals Laboratory and the Medical Devices and Research Institute investigating the underlying neurological mechanisms of audio-visual speech and of various neurological diseases utilising electroencephalograph.